\documentclass{article}
\usepackage{iclr2027,times}

\usepackage{graphicx}
\usepackage{hyperref}
\usepackage{url}
\usepackage{booktabs}
\usepackage{array}
\usepackage{tabularx}
\usepackage{multirow}
\usepackage[table]{xcolor}
\usepackage{amsmath}
\usepackage{amssymb}
\usepackage{amsthm}
\usepackage{useful_math}
\usepackage{algorithm}
\usepackage{wrapfig}
\usepackage{listings}
\usepackage{caption}
\usepackage{cleveref}
\usepackage{twemojis}
\definecolor{gray0}{HTML}{f8f9fa}
\definecolor{gray1}{HTML}{f1f3f5}
\definecolor{gray2}{HTML}{e9ecef}
\definecolor{gray3}{HTML}{dee2e6}
\definecolor{gray4}{HTML}{ced4da}
\definecolor{gray5}{HTML}{adb5bd}
\definecolor{gray6}{HTML}{868e96}
\definecolor{gray7}{HTML}{495057}
\definecolor{gray8}{HTML}{343a40}
\definecolor{gray9}{HTML}{212529}

\definecolor{purple0}{HTML}{e8eaf6}
\definecolor{purple1}{HTML}{c5cae9}
\definecolor{purple2}{HTML}{9fa8da}
\definecolor{purple3}{HTML}{7986cb}
\definecolor{purple4}{HTML}{5c6bc0}
\definecolor{purple5}{HTML}{3f51b5}
\definecolor{purple6}{HTML}{3949ab}
\definecolor{purple7}{HTML}{303f9f}
\definecolor{purple8}{HTML}{283593}
\definecolor{purple9}{HTML}{1a237e}

\definecolor{orange0}{HTML}{fff3e0}
\definecolor{orange1}{HTML}{ffe0b2}
\definecolor{orange2}{HTML}{ffcc80}
\definecolor{orange3}{HTML}{ffb74d}
\definecolor{orange4}{HTML}{ffa726}
\definecolor{orange5}{HTML}{ff9800}
\definecolor{orange6}{HTML}{fb8c00}
\definecolor{orange7}{HTML}{f57c00}
\definecolor{orange8}{HTML}{ef6c00}
\definecolor{orange9}{HTML}{e65100}

\definecolor{teal0}{HTML}{e0f2f1}
\definecolor{teal1}{HTML}{b2dfdb}
\definecolor{teal2}{HTML}{80cbc4}
\definecolor{teal3}{HTML}{4db6ac}
\definecolor{teal4}{HTML}{26a69a}
\definecolor{teal5}{HTML}{009688}
\definecolor{teal6}{HTML}{00897b}
\definecolor{teal7}{HTML}{00796b}
\definecolor{teal8}{HTML}{00695c}
\definecolor{teal9}{HTML}{004d40}

\hypersetup{
  pdfborder={0 0 0},
  colorlinks=true,
  linkcolor=purple6,
  urlcolor=purple6,
  citecolor=gray6,
}

\newcommand{\base}[1]{\raisebox{0pt}[0pt][0pt]{\setlength{\fboxsep}{0.0pt}\colorbox{purple1}{\strut #1}}}
\newcommand{\ours}[1]{\raisebox{0pt}[0pt][0pt]{\setlength{\fboxsep}{0.0pt}\colorbox{orange1}{\strut #1}}}
\newcommand{\shaded}[1]{\raisebox{0pt}[0pt][0pt]{\setlength{\fboxsep}{0.0pt}\colorbox{gray1}{\strut #1}}}
\newcommand{\first}[1]{\cellcolor{teal2}#1}
\newcommand{\second}[1]{\cellcolor{teal1}#1}
\newcommand{\third}[1]{\cellcolor{teal0}#1}
\newcommand{\swatch}[1]{\raisebox{0pt}[0pt][0pt]{\setlength{\fboxsep}{1pt}\colorbox{#1}{\strut\phantom{00}}}}
\usepackage{mathtools}
\newlength{\cbracegap}
\newlength{\cbracepad}
\newcommand{\cbrace}[3]{{\color{#1}\kern-\cbracepad\underbracket[0.6pt][0pt]{\kern\cbracepad{\normalcolor #2}\vphantom{\bigl(}\rule[-\cbracegap]{0pt}{0pt}\kern\cbracepad}_{\mathrm{#3}}\kern-\cbracepad}}
\newcommand{\recbrace}[2]{\cbrace{purple4}{#1}{#2}}
\newcommand{\genbrace}[2]{\cbrace{orange4}{#1}{#2}}

\usepackage{etoolbox}
\makeatletter
\AfterEndEnvironment{algorithm}{\let\@algcomment\relax}
\AtEndEnvironment{algorithm}{\kern2pt\hrule\relax\vskip2pt\@algcomment}
\let\@algcomment\relax
\newcommand\algcomment[1]{\def\@algcomment{\footnotesize#1}}
\renewcommand\fs@ruled{\def\@fs@cfont{\bfseries}\let\@fs@capt\floatc@ruled
  \def\@fs@pre{\vspace{0.0em}\hrule height.7pt depth0pt \kern3pt}%
  \def\@fs@post{\vspace{0.0em}}
  \def\@fs@mid{\kern3pt\hrule\kern2pt}%
  \let\@fs@iftopcapt\iftrue}
\makeatother

\usepackage{aliascnt}

\newaliascnt{proposition}{theorem}

\aliascntresetthe{proposition}
\newaliascnt{lemma}{theorem}

\aliascntresetthe{lemma}
\newaliascnt{corollary}{theorem}

\aliascntresetthe{corollary}

\crefname{figure}{Fig.}{Figs.}       \Crefname{figure}{Figure}{Figures}
\crefname{equation}{Eq.}{Eqs.}       \Crefname{equation}{Equation}{Equations}
\crefname{table}{Tab.}{Tabs.}        \Crefname{table}{Table}{Tables}
\crefname{section}{Sec.}{Secs.}      \Crefname{section}{Section}{Sections}
\crefname{algorithm}{Alg.}{Algs.}    \Crefname{algorithm}{Algorithm}{Algorithms}
\crefname{appendix}{App.}{Apps.}     \Crefname{appendix}{Appendix}{Appendices}
\crefname{proposition}{Prop.}{Props.}  \Crefname{proposition}{Proposition}{Propositions}
\crefname{corollary}{Cor.}{Cors.}      \Crefname{corollary}{Corollary}{Corollaries}
\crefname{lemma}{Lem.}{Lems.}          \Crefname{lemma}{Lemma}{Lemmas}
\newcommand{\abstractheadfrac}{0.165}
\newcommand{\abstractgapfrac}{0.015}
\newlength{\abstractheadwidth}
\newlength{\abstractgap}
\newlength{\abstractbodywidth}
\renewenvironment{abstract}{%
  \setlength{\abstractheadwidth}{\abstractheadfrac\linewidth}%
  \setlength{\abstractgap}{\abstractgapfrac\linewidth}%
  \setlength{\abstractbodywidth}{\dimexpr\linewidth-\abstractheadwidth-\abstractgap\relax}%
  \vskip-0.6em
  \noindent
  \begin{minipage}[t]{\abstractheadwidth}%
    \raggedright\large\sc Abstract\strut
  \end{minipage}%
  \hspace{\abstractgap}%
  \begin{minipage}[t]{\abstractbodywidth}%
    \strut\ignorespaces
    }{%
  \end{minipage}\par\vskip 1ex
}

\title{Sphere Encoder 2}

\author{
   \hspace{-1.0em}
   \small
   Kaiyu Yue$^1$ \quad
   Sean McLeish$^1$ \quad
   Ruchit Rawal$^1$ \quad
   Brian Bartoldson$^2$ \quad
   Menglin Jia$^3$ \quad
   Tom Goldstein$^1$
   \\[5pt]
   \hspace{-0.8em}
   $^1$University of Maryland \quad
   $^2$Lawrence Livermore National Laboratory \quad
   $^3$Cornell University
}

\iclrfinalcopy 
\begin{document}

\maketitle

\begin{abstract}
   Sphere Encoder is an autoencoder that generates images by decoding random points from a high-dimensional latent sphere.
   We identify two limitations of the original formulation that reduce its generation quality.
   First, random points concentrate near the equator relative to the pole on an encoded latent, but the training rotation never reaches this region, leaving a gap that limits one-step generation.
   Second, training for generation with pixel-wise reconstruction loss encourages the decoder to average over plausible images, producing blurry images that lack high-frequency details.
   We present Sphere Encoder 2 to address both limitations, substantially improving generation quality while maintaining the speed and simplicity of the autoencoder.
   Models are released at \href{https://github.com/kaiyuyue/sphere2}{github.com/kaiyuyue/sphere2}.
\end{abstract}

\begin{figure}[H]
   \centering
   \includegraphics[width=1.0\linewidth]{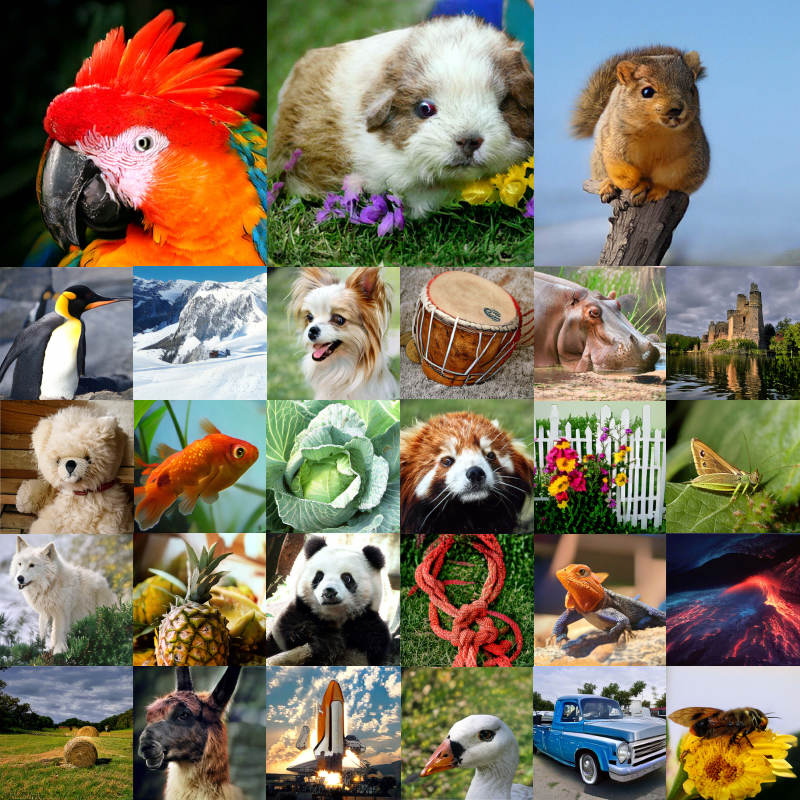}
   \caption{
      \textbf{Curated 1-step generation} without CFG on ImageNet $512 \times 512$ by Sphere Encoder 2.
   }
   \label{fig:imagenet_512px_1step}
\end{figure}

\begin{figure}[t]
   \vspace{-1.0em}
   \centering
   \includegraphics[width=1.0\linewidth]{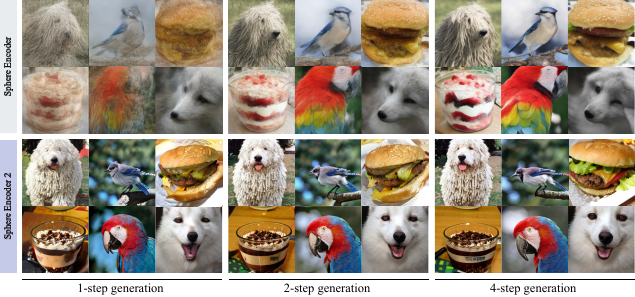}
   \caption{
      \textbf{Sphere Encoder vs. Sphere Encoder 2.}
      Uncurated generation on ImageNet $256 \times 256$ from Sphere Encoder \citep{yue2026image} (top) and Sphere Encoder 2 (bottom) with same classes.
   }
   \label{fig:1_vs_2}
\end{figure}

\section{Introduction}

Sphere Encoders \citep{yue2026image,do2026efficient,man2026sp} are simple standalone autoencoders with strong generative capabilities. They consist of an encoder that maps images into a spherical latent space, and a decoder that maps a latent vector back to an image. When trained properly, the training images cover the latent sphere with an approximately uniform distribution, enabling generation of novel images by sampling a random latent vector and passing it through the decoder. Unlike the many-step generation process used by diffusion models, sphere encoders enable fast image generation in as little as one step.

The original Sphere Encoder \citep{yue2026image} is able to represent the ImageNet distribution, but its single-step generations are blurry (\cref{fig:1_vs_2}, top). Sharp images were obtained by iteratively passing a generated image through the encoder and decoder as many as 4 times, yet soft edges remain even after this refinement.
We trace this to two causes.
First, while noise is added to latent vectors during training to achieve good coverage of the sphere, the original Sphere Encoder only adds up to $85^\circ$ of noise to latents, possibly leaving some of the sphere unseen during training.
Second, it trains the decoder for generation with the pixel-wise reconstruction loss, which is minimized by averaging over all plausible images and thus yields blurry outputs.

Sphere Encoder 2 fixes both issues (\cref{fig:1_vs_2}, bottom). First, we adopt a parameterization of noise that robustly adds a full $90^\circ$ of noise to latents during training, allowing the decoder to see more of the sphere during training. Second, we decouple the decoder's task into two regimes: low-noise reconstruction, which keeps the pixel-wise loss, and high-noise generation, which drops it. In the high-noise/generative regime, we propose a new and efficient latent score matching loss. This encourages outputs to lie on the image manifold without requiring adversarial training. The result is sharp one-step generation on ImageNet $512 \times 512$ (\cref{fig:imagenet_512px_1step}).

Among pixel-space generators on ImageNet, Sphere Encoder 2 achieves the second best FDr$^6$ at $256 \times 256$ and the third best at $512 \times 512$ with only $4 \times 2$ network evaluations (NFE), whereas pixel diffusion models such as PixelDiT \citep{yu2026pixeldit} and PixNerd \citep{wang2026pixnerd} take 200.
This amounts to $28\times$ to $33\times$ less sampling compute in total GFLOPs.
Without classifier-free guidance (CFG) \citep{ho2022classifier}, 4 steps outperform JiT-H \citep{li2026back} and one-step pMF-H \citep{lu2026one} at both resolutions.

\section{Related Work}
\label{sec:related_work}

\textbf{Autoencoders}
\citep{bourlard1988auto,hinton1993autoencoders,kingma2013auto} compress images into a latent space, but making them generate on their own is hard on complex datasets, \eg, ImageNet \citep{deng2009imagenet}.
Hierarchical VAEs such as NVAE \citep{vahdat2020nvae} and VDVAE \citep{child2020very} deepen the latent space, yet their generation quality stays far below diffusion models \citep{dhariwal2021diffusion} and GANs \citep{sauer2023stylegan}.
Deep VAEs \citep{luhman2023high} reach high fidelity on ImageNet only when trained inside the latent space of a pretrained tokenizer.
Autoencoders then settled for the two-stage paradigm: an autoencoder only compresses, and a secondary latent diffusion \citep{rombach2022high} or autoregressive (AR) prior \citep{tian2024visual} generates.
Sphere Encoder \citep{yue2026image} revisits single-stage paradigm with a non-variational autoencoder and reaches competitive quality on ImageNet, but its generation is blurry because it treats generation as reconstruction.
Sphere Encoder 2 still stays single-stage but removes the blur.

\textbf{Sampling from a prior}
is what turns an autoencoder into a generator, and most work focuses on learning a richer prior, from mixture priors \citep{tomczak2018vae} to the hierarchical priors above.
Spherical latents have also been explored, but either at small scale with a von Mises-Fisher posterior \citep{davidson2018hyperspherical}, or with a prior still learned on top of spherical latents, autoregressive in SphereAR \citep{ke2025hyperspherical} and Riemannian flow matching over representation-encoder latents in RJF \citep{kumar2026learning}.
Sphere Latent Encoder \citep{do2026efficient} builds on Sphere Encoder without a learned prior, but it is two-stage again and relies on a RAE decoder \citep{zheng2026diffusion} to reach pixels.
StyleGAN \citep{karras2019style} normalizes its input noise onto a sphere, but it does not synthesize from the sphere directly: a learned mapping network first transforms random inputs into the intermediate latent space $\mathcal{W}$, whose distribution is also a learned prior.
Closest to our setting, \citet{zhao2019latent} decode random points on a sphere directly to pixels, but only working on MNIST and CelebA.
Sphere Encoder is the first to scale sphere-to-pixel decoding to ImageNet: the decoder is trained with noise to cover the entire sphere, so no learned prior is needed.

\textbf{One-/few-step generation}
has two main paths.
Distillation compresses a multi-step teacher into a few steps, by progressively halving steps \citep{salimans2022progressive,meng2023distillation,luhman2021knowledge}, straightening trajectories \citep{liu2022flow,liu2023instaflow}, enforcing consistency \citep{song2023consistency,kim2024consistency}, or matching scores or distributions \citep{luo2023diff,yin2024one,yin2024improved,zhou2024score}, optionally with an adversarial term \citep{sauer2024adversarial}, while from-scratch methods train one/few-step models directly, including consistency training \citep{song2023consistency,song2024improved,geng2025consistency,lu2025simplifying,luo2023latent}, flow-map methods such as shortcut models, IMM, MeanFlow and its variants \citep{frans2025one,zhou2025inductive,geng2025mean,geng2026improved,lu2026one,boffi2024flow}, drifting models \citep{deng2026generative}, GANs \citep{goodfellow2020generative,brock2018large,kang2023scaling,huang2024gan,sauer2023stylegan,hyun2025scalable}, and Sphere Encoders \citep{yue2026image,do2026efficient,man2026sp}.
Sphere Encoder 2 is trained from scratch without distillation.
One step decodes random points on the sphere directly into pixels, and each extra step loops the model by encoding, rotating, and decoding.

\textbf{Distribution matching}
is important for high-fidelity synthesis.
The most effective method is the adversarial loss: a discriminator learns to tell real images from generated ones \citep{goodfellow2020generative,brock2018large,kang2023scaling}.
However, adversarial training needs careful tuning and architecture design \citep{huang2024gan,hyun2025scalable}, and its stability is a problem: BigGAN \citep{brock2018large} collapses after a few hundred ImageNet epochs and must be stopped early, and the discriminator overfits as training proceeds \citep{karras2020training}.
Distribution matching distillation (DMD) \citep{yin2024one} proposes an alternative: a pretrained diffusion teacher provides the real score, a second model tracks the fake score online, and their difference moves samples toward the data distribution \citep{wang2023prolificdreamer,zhou2024score}.
The cost is the teacher: DMD requires a pretrained diffusion model of the same data, so it is a distillation method rather than a standalone loss for any generator.
Without a teacher, SD flow \citep{weber2023score} shows that the difference of noise-smoothed real and fake scores is the flow that optimally reduces their KL divergence, but only on toy data.
SMT \citep{jayashankar2025score} learns the score of the real/fake mixture online, but only at $64 \times 64$ resolution.
Concurrent work matches real and generated features of frozen encoders with MMD or optimal transport \citep{feng2026representation,zhang2026unifying}, estimating the transport direction non-parametrically from a batch, but so far only as post-training of pretrained multi-step generators.
In this work, we propose a lightweight score matching loss without a pretrained teacher.
It trains two score models online from scratch, one for real and one for fake features, each a two-layer transformer following standard EDM \citep{karras2022elucidating}, and both operate in a frozen feature space rather than pixel space, as in Projected GAN \citep{sauer2021projected}, StyleGAN-T \citep{sauer2023stylegan}, and drifting loss \citep{deng2026generative}.
Like a GAN loss, the score matching loss depends only on real and generated images, not on the generator itself, so in principle it can be plugged into any generator training pipeline.
Its score models can also be conditioned on class or caption embeddings.

\section{Preliminaries: Sphere Encoder}

Sphere Encoder \citep{yue2026image} is a non-variational autoencoder that can be efficiently sampled to generate images.
It consists of an encoder $\cE$ and a decoder $\cD$.
Given an input image $\X$\footnote{
   Bold capital letters denote matrices, and bold lower-case letters denote column vectors.
   The notation $\x_i$ represents the $i^{\text{th}}$ column of a matrix $\X$, while
   $x_{ij}$ denotes its entry in the $i^{\text{th}}$ row and $j^{\text{th}}$ column.
   Non-bold letters denote scalars, and $\vect(\X)$ denotes the vectorization of $\X$.
}
$\in \real^{H \times W \times 3}$, the encoder produces a latent $\Z = \cE(\X) \in \real^{N \times D}$, where $N$ is the number of tokens and $D$ is the feature dimension.
The latent $\Z$ is then projected onto a global sphere by the spherify function $\cF$:
\begin{align}
   \label{eq:spherify}
   \z
   = \cF(\Z)
   = \sqrt{d}\,
   \frac{\vect(\Z) \ }{\|\vect(\Z)\|_2}
   \in \sqrt{d} \ \mathbb{S}^{d-1},
   \quad d = N \times D,
\end{align}
where $\mathbb{S}^{d-1}$ denotes the unit sphere in $\real^{d}$, so
$\z$ lies on a sphere of radius $\sqrt{d}$.
The decoder then maps the spherical latent back to the image space: $\hat{\X} = \cD(\z)$, reconstructing the original image.

For generation, the Sphere Encoder exploits the fact that the latent space is approximately uniformly distributed on the sphere, enabling the decoder to be run with a uniform sample as input.
A latent vector is chosen by spherifying a random Gaussian vector, resulting in a uniformly sampled vector on the sphere \citep{blum2020foundations}, and the decoder transforms it into an image:
\begin{align}
   \label{eq:generation}
   \hat{\X}
   = \cD \bigl( \cF(\be) \bigr),
   \quad
   \be \sim \cN(\zero, \I_{d}).
\end{align}

\subsection{Gap to the Equator}

For the direct pixel generation in \cref{eq:generation} to succeed, the decoder must handle latents that are far from any encoded $\z$ seen at training time.
Sphere Encoder bridges the gap between the discrete training distribution and continuous uniform distribution using a process called \emph{spherify with noise}, which adds noise to a clean latent $\z$ before train-time reconstruction:
\begin{align}
   \label{eq:spherify_with_noise}
   \hat{\z}(\alpha)
   = \cF\left(\z + \tan(\alpha) \ \cF(\be) \right),
\end{align}
where $\alpha$ approximates the angle between $\z$ and $\hat{\z}(\alpha)$ in high dimensions.
During training, $\alpha$ is uniformly sampled from $[0, \alpha_\mathrm{max}]$ with $\alpha_\mathrm{max} = 85^\circ$.

This objective leaves a mismatch between the latents seen in training and those used for generation.
Specifically, on a high-dimensional sphere, almost all of the surface measure concentrates in a thin band around the equator relative to any fixed direction \citep{blum2020foundations,goldstein2018phasemax}.
Relative to $\z$, a random point $\cF(\be)$ lies near the equator at $\alpha \rightarrow 90^\circ$.
With $\alpha_\mathrm{max} = 85^\circ$, the decoder is never trained on the equatorial band, and the residual gap of $5^\circ$ limits its one-step generation quality.
Closing the gap by setting $\alpha_\mathrm{max} = 90^\circ$ is not an option in \cref{eq:spherify_with_noise}, since $\tan(\alpha)$ diverges as $\alpha \rightarrow 90^\circ$.
This main limitation motivates a new formulation that reaches the equator exactly.

\section{Sphere Encoder 2}
\label{sec:sphere2}

In this section, we introduce Sphere Encoder 2, which improves generation quality by closing the gap to the equator.
We explain its design choices through ablation studies.

\begin{figure}[h]
   \vspace{-1.0em}
   \centering
   \includegraphics[width=0.8\linewidth]{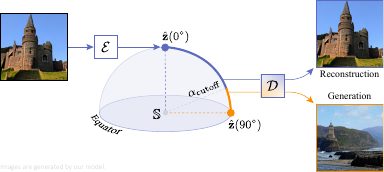}
   \caption{
      \textbf{Sphere Encoder 2.}
      The encoder $\cE$ maps an image to a clean latent $\hat{\z}(0^\circ)$ on the sphere $\mathbb{S}$, which is then rotated toward a random direction by an angle $\alpha \in [0, 90^\circ]$ in \cref{eq:rotation}.
      The cutoff angle $\alpha_\mathrm{cutoff}$ splits the arc into a reconstruction regime ($\alpha \leqslant \alpha_\mathrm{cutoff}$) and a generation regime ($\alpha \geqslant \alpha_\mathrm{cutoff}$).
      The decoder $\cD$ serves both: near the pole it reconstructs the input, and at the equator $\hat{\z}(90^\circ) = \cF(\be)$ it generates a new image from a random latent in \cref{eq:generation}.
   }
   \label{fig:model}
\end{figure}

\subsection{Rotation to the Equator}

We replace additive perturbation and reprojection in \cref{eq:spherify_with_noise} with an explicit rotation:
\begin{align}
   \label{eq:rotation}
   \hat{\z}(\alpha)
   = \cos(\alpha) \ \z + \sin(\alpha) \ \cF(\be),
   \quad \alpha \in [0, 90^\circ].
\end{align}
Since $\z$ and $\cF(\be)$ are approximately orthogonal in high dimensions, $\hat{\z}(\alpha)$ stays on the sphere of radius $\sqrt{d}$, and $\alpha$ is exactly the angle of rotation away from $\z$ toward $\cF(\be)$.
The endpoints are well defined: $\hat{\z}(0^\circ) = \z$ recovers the clean latent, and $\hat{\z}(90^\circ) = \cF(\be)$ lands on the equator (see \cref{fig:model}).

This objective closes the gap in two ways.
First, the range of $\alpha$ now includes $90^\circ$, so the decoder is trained on the equatorial band that one-step generation samples from.
Second, although the distribution of the clean latents $\z$ is not necessarily uniform, $\hat{\z}(90^\circ)$ is uniform on the equatorial subsphere relative to $\z$, which is exactly the distribution of the generation latents $\cF(\be)$ in \cref{eq:generation}.

\subsection{Still an Autoencoder}

Sphere Encoder 2 is a standard autoencoder with a single encoder and decoder.
We adopts a frozen DINOv3-small+ \citep{simeoni2025dinov3} as the encoder $\cE$ for fast forward and high-quality generation.
Each image is encoded into $N = 16 \times 16$ tokens with $D = 384$ features per token, giving a latent size of $d = 98{,}304$.
The latent size is shared across image resolutions of $256 \times 256$ and $512 \times 512$, with patch sizes of $16 \times 16$ and $32 \times 32$ in pixel prediction head of the decoder, respectively.

Only the decoder $\cD$, just a ViT \citep{dosovitskiy2020image}, is trained and scaled.
We name models by decoder size, Sphere2-B and Sphere2-L for base and large decoders.
The decoder is conditioned on the class embedding $\y$ and the rotation angle $\alpha$, so it can behave differently along the rotation arc.
Formally, the unconditional generation \cref{eq:generation} becomes
\begin{align}
   \label{eq:generation_conditional}
   \hat{\X} = \cD(\cF(\be), \y, \alpha).
\end{align}

We embed $\alpha$ in the same way as diffusion timesteps \citep{nichol2021glide,nichol2021improved}.
The angle embedding is added to both the positional embedding of the input tokens and the class embedding used by adaLN-Zero \citep{peebles2023scalable,perez2018film}.

\noindent
\begin{minipage}[c]{0.41\linewidth}
   \cref{alg:forward} summarizes the sampling process.
   One-step generation decodes a random point on the sphere.
   The few-step generation repeatedly encodes the generated image, rotates its latent toward the fixed noise direction $\cF(\be)$ by a fixed angle $\alpha$, and decode it again.
   Unless otherwise specified, ablations use $\alpha = 85^\circ$ and no classifier-free guidance (CFG) \citep{ho2022classifier}; the main results use an optimal per-model angle (\cref{appsec:abl_sampling_angle}).
\end{minipage}
\hfill
\begin{minipage}[c]{0.56\linewidth}
   \vspace{-1.0em}
   \begin{algorithm}[H]
      \caption{Sampling forward of Sphere Encoder 2.}
      \label{alg:forward}
      \lstset{
         backgroundcolor=\color{white},
         basicstyle=\fontsize{8.0pt}{7.4pt}\ttfamily\selectfont,
         columns=fullflexible,
         breaklines=true,
         captionpos=b,
         commentstyle=\fontsize{7.5pt}{7.5pt}\color{purple4},
         keywordstyle=\fontsize{7.5pt}{7.5pt},
         aboveskip=0.4em,
         belowskip=0.4em,
      }
      \begin{lstlisting}[language=python]
e = Normal(0, 1).sample([d])  # random point

# one-step generation
x = D(F(e), y, alpha)

# few-step refinement
for _ in range(K - 1):  # K steps in total
   z = cos(alpha) * F(E(x)) + sin(alpha) * F(e)
   x = D(z, y, alpha)

return x
         \end{lstlisting}
   \end{algorithm}
\end{minipage}

\subsection{Angle Cutoff: Two Regimes on One Sphere}

Sphere Encoder treats generation as reconstruction: the pixel-wise reconstruction loss is applied to the rotated latent $\hat{\z}(\alpha)$ along the whole arc.
A pixel-space loss, however, is minimized by the mean of all plausible images, so the decoder learns to generate mean images in one-step generation.

To address this issue, we first split the rotation arc at a cutoff $\alpha_\mathrm{cutoff}$ into two regimes, each with its own training losses.
The reconstruction regime covers the low angles, $\alpha \in [0, \alpha_\mathrm{cutoff}]$, where the decoder is trained with the pixel-wise reconstruction loss:
\begin{align}
   \cL_\mathrm{pix\text{-}recon} = \cL_\mathrm{LPIPS}(\hat{\X}, \X) + \cL_\mathrm{L1}(\hat{\X}, \X),
\end{align}
where $\cL_\mathrm{LPIPS}$ is the perceptual loss \citep{zhang2018unreasonable}.
This regime keeps the clean latent $\hat{\z}(0^\circ)$ faithful, and anchors the low-angle/noise area so that rotating toward the equator starts from a semantically meaningful latent.

\noindent
\begin{minipage}[c]{0.56\linewidth}
   The generation regime covers angles $\alpha \in [\alpha_\mathrm{cutoff}, 90^\circ]$ near the equator.
   The generated images from this regime should stay semantically consistent with the input image, yet vary freely in appearance and structure.
   Before introducing the generation loss, we define a smooth cutoff function to weight the losses across these two regimes.
   The loss weight $w$ for the generation regime is a standard sigmoid centered at $\alpha_\mathrm{cutoff}$ (\cref{fig:cutoff_weight}):
   \begin{align}
      w(\alpha; \alpha_\mathrm{cutoff}, \beta) = \mathrm{sigmoid}\left(\frac{\alpha-\alpha_\mathrm{cutoff}}{\beta/4}\right).
   \end{align}
   The reconstruction regime receives $1 - w$.
   The bandwidth $\beta > 0$ controls the transition width: the scale $\beta / 4$ makes $w$ rise from $0.02$ to $0.98$ within $\pm\beta$ of the cutoff.
\end{minipage}
\hfill
\begin{minipage}[c]{0.40\linewidth}
   \centering
   \includegraphics[width=\linewidth]{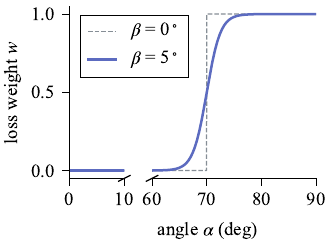}
   \vspace{-1.5em}
   \captionof{figure}{
      \textbf{Loss weight $w$} for generation regime with $\alpha_\mathrm{cutoff} = 70^\circ$ and $\beta = 5^\circ$ used by default.
   }
   \label{fig:cutoff_weight}
\end{minipage}

\subsection{Taming the Generation Regime}

Reconstruction has been well studied in prior work \citep{yao2025reconstruction,zheng2026diffusion,esser2024scaling,yue2026image}, and is not the focus of this work.
We keep the reconstruction regime as is, without adversarial sharpening, so that any change in generation quality can be attributed to the generation losses alone.
Our focus in this work is the generation regime, with the goal set above: along the rotation arc in \cref{fig:model}, generated images should stay semantically consistent with the reconstruction while varying in appearance and structure.

\subsubsection{Semantic Alignment}
\label{sec:sem_align}

To keep the generation $\hat{\X}$ semantically consistent with the input $\X$, we match their LPIPS ConvNeXt logits \citep{zhang2018unreasonable} with a centered cosine similarity:
\begin{align}
   \cL_\mathrm{sem} = \cL_\mathrm{LPIPS\text{-}logits}(\hat{\X}, \X).
\end{align}
This loss adds no cost: it shares the LPIPS perceptual loss forward in the reconstruction regime.
It requires no labels, since the target is the input image itself rather than a class or a caption.
In principle, a different frozen pretrained network with a global semantic embedding could also be used, such as the CLS token of CLIP \citep{radford2021learning}, or SigLIP \citep{tschannen2025siglip}.

\begin{figure}[h]
   \centering
   \includegraphics[width=1.0\linewidth]{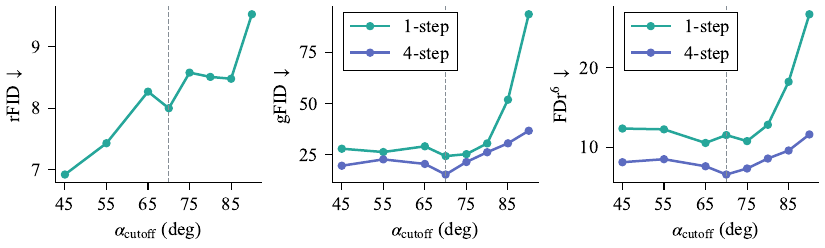}
   \caption{
      \textbf{Cutoff angle $\alpha_\mathrm{cutoff}$}, which splits the rotation arc into the reconstruction and generation regimes.
      Generation improves as the cutoff moves toward the equator until $70^\circ$.
      We omit $\alpha_\mathrm{cutoff}=10^\circ$ from the plot: it lies far outside the sweep and discards most of the signal, so generation does not recover with more steps (FDr$^6$ 37.1 and gFID 77.7 at 1 step, 30.2 and 64.0 at 4 steps).
   }
   \label{fig:abl_cutoff_res}
\end{figure}

The cutoff angle $\alpha_\mathrm{cutoff}$ decides how much of the arc is devoted to generation.
Thus how far generated images may deviate from the input under the semantic alignment loss.
We ablate this trade-off with a Sphere2-B trained on Oxford Flowers $256\times256$ \citep{nilsback2008automated} in \cref{fig:abl_cutoff_res}.

First, in \cref{fig:abl_cutoff_res} left, reconstruction quality degrades as the cutoff moves past $45^\circ$ toward the equator.
Increasing the cutoff extends reconstruction supervision to more strongly corrupted training inputs and changes its balance with the generation losses.
Second, \cref{fig:abl_cutoff_res} middle and right show that 1-step and 4-step generation quality (measured by FID and FDr$^6$, respectively) improves as $\alpha_\mathrm{cutoff}$ moves toward the equator, peaking at $70^\circ$.
We use $\alpha_\mathrm{cutoff} = 70^\circ$ for all experiments by default.

\begin{figure}[h]
   \centering
   \includegraphics[width=1.0\linewidth]{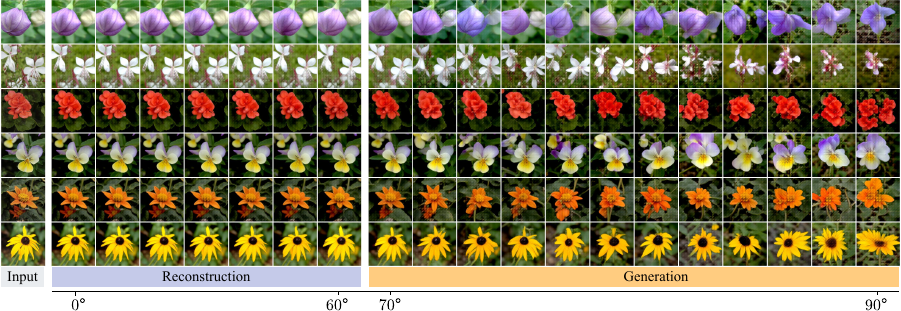}
   \caption{
      \textbf{Reconstruction and generation regimes} at different angles from $0^\circ$ to $90^\circ$ on Oxford Flowers $256\times256$.
      On the same rotation arc with $\alpha_\mathrm{cutoff} = 70^\circ$, the decoder reconstructs the input below the cutoff and generates new images above it.
   }
   \label{fig:abl_cutoff_vis_flowers}
\end{figure}

\cref{fig:abl_cutoff_vis_flowers} visualizes the decoder outputs along the rotation arc, given the input image on the left.
Below $\alpha_\mathrm{cutoff} = 70^\circ$, the decoder reconstructs the input with minor variations; the softness is the intended effect of omitting adversarial training.
Beyond the cutoff, generated images remain semantically consistent with the input while differing in structure and style, which is exactly the behavior the generation regime is designed for.
The deviation grows with the angle, and at $90^\circ$ the latent is pure noise and carries no information about the input, so the output depends only on the class.
Current generated images are waffle-ish because the semantic alignment constrains what the decoder generates, but nothing yet drives it toward sharp, high-fidelity textures.
We address this next.

\subsubsection{Fidelity Improvement}

Improving synthesis fidelity with a loss is a long-standing problem in generative modeling.
The adversarial loss \citep{goodfellow2020generative,brock2018large,kang2023scaling} remains the most effective approach, but it requires careful tuning and architecture design \citep{huang2024gan,hyun2025scalable}, and its long training suffers from stability issues \citep{karras2020training}.

We seek a simple alternative with two properties: (1) it is stable over long training on large datasets, \eg, hundreds of epochs on ImageNet; and (2) it consistently yields effective fidelity supervision as the generator improves, rather than saturating once the generator catches up.
Inspired by DMD \citep{yin2024one} and drifting loss \citep{deng2026generative}, we propose a score matching loss that is agnostic to the generator and plugs in as a standalone loss module, in the same way as a GAN loss.

\begin{figure}[h]
   \centering
   \includegraphics[width=0.8\linewidth]{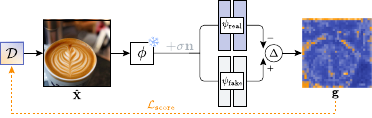}
   \caption{
      \textbf{Score matching loss.}
      Fake features from the frozen extractor $\phi$ are denoised by the real and fake score models (each a two-layer transformer EDM denoiser \citep{karras2022elucidating}), and their difference gives the direction $\g$ that pulls fake features toward real ones.
      During transport, both score models are frozen and run once each; only the decoder $\cD$ is updated by $\g$ through $\cL_{\text{score}}$.
   }
   \label{fig:score_matching}
\end{figure}

Like a GAN loss, it trains an auxiliary model online alongside the generator, but instead of a discriminator that classifies real from fake, it trains two lightweight score models, each only two transformer layers following standard EDM \citep{karras2022elucidating}, that estimate the real and fake distributions in a frozen feature space; their difference gives the direction to move fake features toward real ones.
Like a discriminator \citep{reed2016generative,sauer2023stylegan,zhang2018stackgan,kang2023scaling,sauer2024adversarial}, the score models can be unconditional or conditioned on the class or caption.

We extract features of the input image $\X$ (real) and the generated image $\hat{\X}$ (fake) from a frozen pretrained network $\phi(\cdot)$, such as ConvNeXt \citep{woo2023convnext} or DINO \citep{caron2021emerging}:
\begin{align}
   \label{eq:score_matching_clean_feature}
   \bv_j = \frac{\phi_j(\X) - \mu_j}{s_j},
   \qquad
   \hat{\bv}_j = \frac{\phi_j(\hat{\X}) - \mu_j}{s_j},
\end{align}
where $j$ indicates the $j^{\mathrm{th}}$ layer of $\phi$.
$\bv_j, \hat{\bv}_j \in \real^{L_j \times C_j}$ have $L_j$ tokens and $C_j$ channels.
$\mu_j$ and $s_j$, estimated on real features only, standardize $j^{\mathrm{th}}$-layer features to zero mean and standard deviation $\sigma_\mathrm{data} = 0.5$, as assumed by EDM (see \cref{appsec:score_matching}).
We drop $j$ for simplicity in the following discussion.

We build two score models, $\psi_\mathrm{real}$ and $\psi_\mathrm{fake}$, trained on the real features $\bv$ and the fake features $\hat{\bv}$, respectively.
Both follow EDM \citep{karras2022elucidating} with its standard preconditioning, and each has only two transformer layers.
For a feature $\bu \in \{\bv, \hat{\bv}\}$ and a noise level $\sigma$, the noisy feature is
\begin{align}
   \label{eq:score_matching_noisy_feature}
   \bu_\sigma = \bu + \sigma \n,
   \quad
   \n \sim \cN(\zero, \I),
\end{align}
then $\psi_\mathrm{real}$ and $\psi_\mathrm{fake}$ learn to denoise $\bu_\sigma$ with $\E_{\sigma, \n} \big[ \lambda(\sigma) \| \psi(\bu_\sigma, \sigma) - \bu \|_2^2 \big]$, where $\lambda(\sigma) = (\sigma^2 + \sigma_\mathrm{data}^2) / (\sigma\, \sigma_\mathrm{data})^2$ is the EDM loss weighting and $\sigma$ follows the EDM log-normal $p(\sigma)$.

To supervise the generator, we freeze both score models, evaluate them on the noisy fake feature $\hat{\bv}_\sigma$, and take their difference, as illustrated in \cref{fig:score_matching}:
\begin{align}
   \label{eq:score_matching_transport}
   \g = \frac{
      \psi_\mathrm{fake}(\hat{\bv}_\sigma, \sigma) - \psi_\mathrm{real}(\hat{\bv}_\sigma, \sigma)
   }{
      \| \hat{\bv} - \psi_\mathrm{real}(\hat{\bv}_\sigma, \sigma) \|_1 / LC
   },
\end{align}
where the denominator is the denoising residual of the real score model, which normalizes the magnitude of $\g$ across noise levels following DMD \citep{yin2024one}.
$\g$ carries no gradient and is applied to the generator through a proxy loss:
\begin{align}
   \label{eq:score_matching_loss}
   \cL_\mathrm{score} = \frac{1}{2} \| \hat{\bv} - \mathrm{sg}(\hat{\bv} - \g) \|^2_2,
\end{align}
where $\mathrm{sg}(\cdot)$ is the stop-gradient operator.
Its gradient with respect to $\hat{\bv}$ is exactly $\g$, so a step moves the fake features toward the real distribution.
In practice, we score two layers of $\phi$ and balance their losses so that each contributes a gradient of equal magnitude.
By default, we use the frozen ConvNeXt V2-N(ano) \citep{woo2023convnext} (14.98M) as $\phi$ and train the score models on the outputs of its two middle stages, \ie, the second and third of its four stages, giving a module with just 24.16M parameters in total.
All details of the implementation are described in \cref{appsec:score_matching}, and its training dynamics are shown in \cref{appsec:score_dynamics}.

\subsubsection{Latent Consistency}

We also keep the latent consistency loss of Sphere Encoder \citep{yue2026image}, which encodes the generated image and matches its latent to that of the input:
\begin{align}
   \cL_\mathrm{lat\text{-}con} = 1 - \mathrm{cos\text{-}sim}\big(\cE(\hat{\X}), \cE(\X)\big).
\end{align}
Since $\cE$ is the frozen DINOv3-small+, this becomes semantic alignment again, now in the latent space rather than in the logit space of $\cL_\mathrm{sem}$.
It adds one encoder forward on the generated image.

\subsubsection{Training Objective}

In summary, Sphere Encoder 2 trains the decoder $\cD$ with a weighted combination of four losses:
\begin{align}
   \cL_\mathrm{total}
   = \recbrace{(1 - w) \, \cL_\mathrm{pix\text{-}recon}}{reconstruction \, regime}
   + \genbrace{w \, \bigl( \cL_\mathrm{sem} + \cL_\mathrm{score} + 0.1 \, \cL_\mathrm{lat\text{-}con} \bigr)}{generation \, regime}.
\end{align}
\cref{tab:abl_component_roadmap} ablates these components one at a time, and each improves generation at every step count.

\section{Experiments}
\label{sec:experiments}

\cref{tab:fnl_flowers,tab:fnl_imagenet} summarize the results of Sphere Encoder 2 on Oxford Flowers and ImageNet, respectively.
For both datasets, we generate 50K images and report FDr$^{6}$ \citep{yang2026representation} and gFID \citep{heusel2017gans} against the whole training set.
We prioritize FDr$^{6}$ over gFID, since it averages over six feature spaces rather than Inception alone, which provides a more robust evaluation of generation quality.
For Sphere2 models, the sampling angle (\cref{appsec:abl_sampling_angle}) and guidance angle (\cref{appsec:cfg}) are chosen by FDr$^{6}$, and gFID is reported at the same setting rather than tuned separately, since the two metrics prefer different settings.
All training and evaluation details are in \cref{appsec:impl}.

\begin{figure}[h]
   \vspace{-1.0em}
   \centering
   \includegraphics[width=1.0\linewidth]{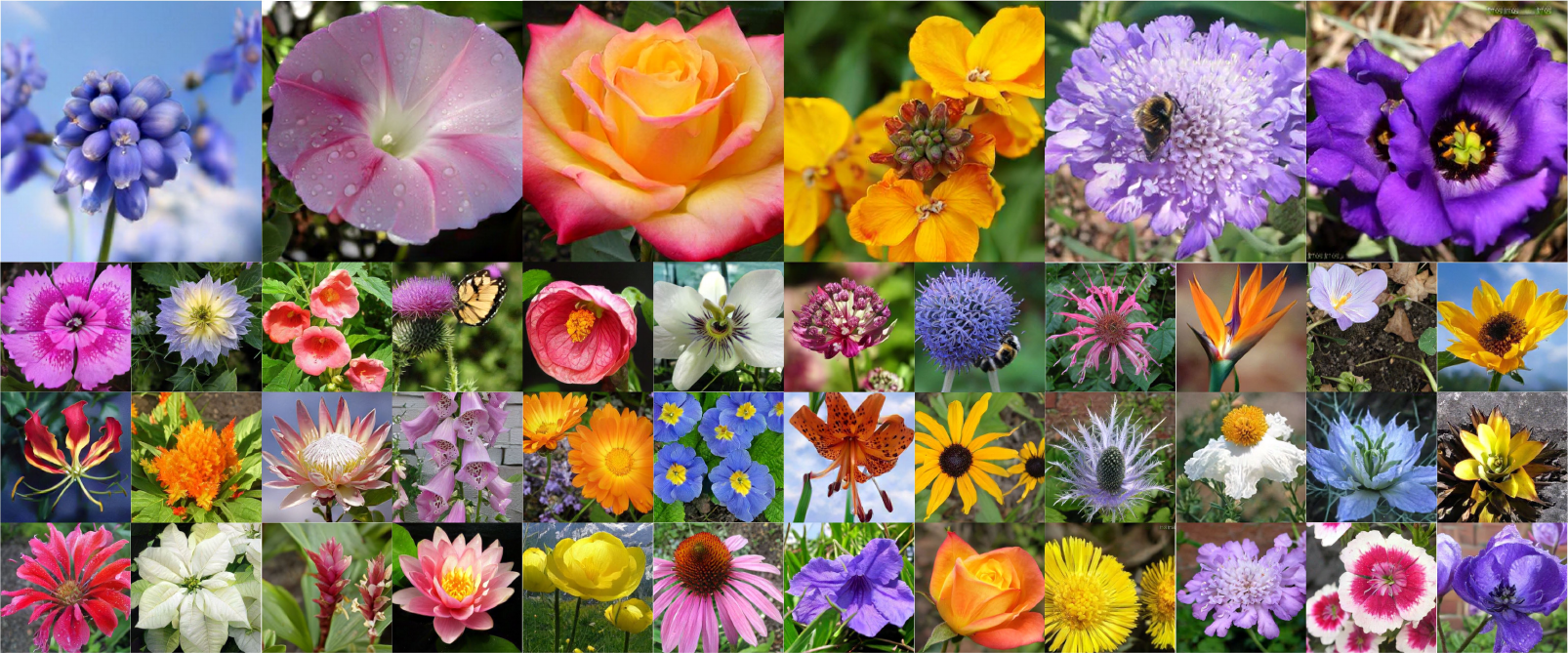}
   \caption{
      \textbf{Uncurated 4-step generation} without CFG on Oxford Flowers $512 \times 512$ by Sphere2-L.
   }
   \label{fig:flowers_512px_nstep}
\end{figure}

On Oxford Flowers $256 \times 256$, Sphere2-B with one step and no CFG outperforms both the previous Sphere Encoder \citep{yue2026image} and Sphere Latent Encoder \citep{do2026efficient} at 4 steps with CFG, while using $6\times$ fewer parameters than Sphere Encoder.
At this resolution, Sphere2-L improves FDr$^{6}$ over Sphere2-B at every step count but has worse gFID, showing that the two metrics are not always correlated.
\cref{fig:flowers_512px_nstep} shows uncurated 4-step samples at $512 \times 512$.

\begin{table}[H]
    \centering
    \caption{
        \textbf{Main comparison on Oxford Flowers}.
        Compared with Sphere Encoder \citep{yue2026image} and Sphere Latent Encoder \citep{do2026efficient}.
        Params for Sphere2 cover both encoder and decoder; for latent-space models they cover the generator only.
        Sphere2 has no CFG in sampling.
    }
    \label{tab:fnl_flowers}
    \small
    \setlength{\tabcolsep}{4.1pt}
    \begin{tabular}{l c c c cccc cccc}
        \toprule
        \multirow{2}{*}[-0.55ex]{Model}                        &
        \multirow{2}{*}[-0.55ex]{Params}                       &
        \multirow{2}{*}[-0.55ex]{Guid.}                        &
        \multirow{2}{*}[-0.55ex]{NFE}                          &
        \multicolumn{4}{c}{FDr$^{6}$ $\downarrow$ @ $K$ steps} &
        \multicolumn{4}{c}{gFID $\downarrow$ @ $K$ steps}                                                                                                              \\
        \cmidrule(lr){5-8} \cmidrule(lr){9-12}
                                                               &      &     &              & 1        & 2    & 4    & 6    & 1        & 2        & 4        & 6        \\
        \midrule
        \rowcolor{gray1}
        \multicolumn{12}{l}{\textit{Image Size $256 \times 256$}}                                                                                                      \\
        Sphere \citeyearpar{yue2026image}                      & 948M & CFG & $K \times 2$ & 10.59    & 5.44 & 3.96 & 3.92 & 25.12    & 14.08    & 11.25    & 10.63    \\
        Sphere Latent \citeyearpar{do2026efficient}            & 130M & CFG & $K \times 2$ & --       & --   & --   & --   & --       & 12.22    & \;\;8.61 & \;\;7.85 \\
        \cmidrule(lr){1-12}
        \textbf{Sphere2-B}                                     & 159M & --  & $K$          & \;\;2.67 & 2.23 & 2.17 & 2.19 & \;\;7.62 & \;\;7.45 & \;\;7.50 & \;\;7.55 \\
        \textbf{Sphere2-L}                                     & 487M & --  & $K$          & \;\;2.19 & 1.85 & 1.82 & 1.83 & \;\;9.36 & \;\;8.33 & \;\;7.98 & \;\;7.93 \\
        \midrule
        \rowcolor{purple0}
        \multicolumn{12}{l}{\textit{Image Size $512 \times 512$}}                                                                                                      \\
        \textbf{Sphere2-B}                                     & 161M & --  & $K$          & \;\;4.86 & 4.18 & 4.08 & 4.10 & \;\;9.47 & \;\;8.83 & \;\;8.49 & \;\;8.32 \\
        \textbf{Sphere2-L}                                     & 489M & --  & $K$          & \;\;4.22 & 3.67 & 3.60 & 3.61 & \;\;9.56 & \;\;8.34 & \;\;7.70 & \;\;7.59 \\
        \bottomrule
    \end{tabular}
\end{table}

\begin{table}[!b]
    \centering
    \caption{
        \textbf{Main comparison on ImageNet} $256 \times 256$ and $512 \times 512$.
        Params and GFLOPs for Sphere2 cover both encoder and decoder; for latent-space models they cover the generator only and \emph{exclude} the VAE decoder.
        GFLOPs are measured for a single forward, so Total GFLOPs $=$ GFLOPs $\times$ NFE.
        Guidance: CFG is classifier-free guidance applied at sampling, so NFE is doubled; AG is AutoGuidance \citep{karras2024guiding}, which doubles NFE with a smaller guiding model; CFG$^\ast$ denotes guidance folded into the training target for single-forward sampling.
        Among pixel-space models, taking each model at its best setting, the best, second and third FDr$^{6}$ per resolution are shaded \swatch{teal2}, \swatch{teal1} and \swatch{teal0}.
        $^\dagger$ marks Sphere2 models trained with the $\cL_\mathrm{FD\text{-}lite}$ loss in \cref{appsec:distribution_shift}.
    }
    \label{tab:fnl_imagenet}
    \small
    \setlength{\tabcolsep}{4.4pt}
    \renewcommand{\arraystretch}{0.95}
    \begin{tabular}{l r c r@{\ $\times$\ }l cc cc}
        \toprule
        \multirow{2}{*}[-0.5ex]{Model}              &
        \multirow{2}{*}[-0.5ex]{Params}             &
        \multirow{2}{*}[-0.5ex]{Guid.}              &
        \multicolumn{2}{c}{Total GFLOPs}            &
        \multicolumn{2}{c}{$256 \times 256$}        &
        \multicolumn{2}{c}{$512 \times 512$}                                                                                                   \\
        \cmidrule(lr){4-5} \cmidrule(lr){6-7} \cmidrule(lr){8-9}
                                                    &                        &                   & GFLOPs                 & NFE
                                                    & FDr$^{6}$ $\downarrow$ & gFID $\downarrow$ & FDr$^{6}$ $\downarrow$ & gFID $\downarrow$  \\
        \midrule
        \rowcolor{gray1}
        \multicolumn{9}{l}{\textit{Latent space}}                                                                                              \\
        SiT-XL/2 \citeyearpar{ma2024sit}
                                                    & 675M                   & CFG               & 119                    & $250 \times 2$
                                                    & \; 8.44                & 2.06              & \; --                  & \ \ 2.62           \\
        SiT-XL/2 + REPA \citeyearpar{yu2024representation}
                                                    & 675M                   & CFG               & 119                    & $250 \times 2$
                                                    & \; 5.45                & 1.42              & \; --                  & \ \ 2.08           \\
        LightningDiT-XL/2 \citeyearpar{yao2025reconstruction}
                                                    & 675M                   & CFG               & 119                    & $250 \times 2$
                                                    & \; 4.57                & 1.35              & \; --                  & --                 \\
        DDT-XL/2 \citeyearpar{wang2026ddt}
                                                    & 675M                   & CFG               & 119                    & $250 \times 2$
                                                    & \; 5.70                & 1.26              & \; --                  & \ \ 1.28           \\
        RAE-DiT$^\text{DH}$-XL/2 \citeyearpar{zheng2026diffusion}
                                                    & 839M                   & \; AG             & 146                    & \; $50 \times 2$
                                                    & \; 3.26                & 1.13              & \; 2.74                & \ \ 1.13           \\
        MAR-L \citeyearpar{li2024autoregressive}
                                                    & 478M                   & CFG               & 129                    & $256 \times 2$
                                                    & \; 6.68                & 1.80              & \; --                  & --                 \\
        MAR-H
                                                    & 942M                   & CFG               & 252                    & $256 \times 2$
                                                    & \; 5.61                & 1.56              & \; --                  & --                 \\
        Sphere Latent-XL \citeyearpar{do2026efficient}
                                                    & 675M                   & CFG               & 230                    & \; \; $6 \times 2$
                                                    & \; 4.08                & 2.11              & \; --                  & --                 \\
        GAT-XL/2 \citeyearpar{hyun2025scalable}     & 602M                   & --                & 119                    & \; \; 1
                                                    & \; 9.26                & 2.06              & \; --                  & --                 \\
        Drift-L \citeyearpar{deng2026generative}
                                                    & 463M                   & \; CFG$^\ast$     & 87                     & \; \; $1$
                                                    & 10.92                  & 1.53              & \; --                  & --                 \\
        iMF-XL \citeyearpar{geng2026improved}
                                                    & 610M                   & \; CFG$^\ast$     & 174                    & \; \; $1$
                                                    & \; 8.39                & 1.82              & \; --                  & --                 \\
        iMF-XL
                                                    & 610M                   & \; CFG$^\ast$     & 174                    & \; \; $2$
                                                    & \; 7.48                & 1.61              & \; --                  & --                 \\
        \midrule
        \rowcolor{purple0}
        \multicolumn{9}{l}{\textit{Pixel space}}                                                                                               \\
        PixNerd-XL \citeyearpar{wang2026pixnerd}
                                                    & 700M                   & CFG               & 130                    & $100 \times 2$
                                                    & \; \third{5.01}        & 2.10              & \; \second{5.68}       & \ \ 2.84           \\
        PixelDiT-XL \citeyearpar{yu2026pixeldit}    & 797M                   & CFG               & 151                    & $100 \times 2$
                                                    & \; \first{4.46}        & 1.66              & \; \first{4.88}        & \ \ 1.84           \\
        JiT-B \citeyearpar{li2026back}
                                                    & 131M                   & CFG               & 25                     & \; $50 \times 4$
                                                    & 15.78                  & 3.66              & 16.31                  & \ \ 4.19           \\
        JiT-L
                                                    & 459M                   & CFG               & 88                     & \; $50 \times 4$
                                                    & 10.73                  & 2.36              & 11.35                  & \ \ 2.53           \\
        JiT-H
                                                    & 953M                   & CFG               & 182                    & \; $50 \times 4$
                                                    & \; 7.66                & 1.86              & \; 8.66                & \ \ 1.94           \\
        Sphere-XL \citeyearpar{yue2026image}
                                                    & 1.3B                   & CFG               & 977                    & \; \; $4 \times 2$
                                                    & 10.44                  & 4.02              & \; --                  & --                 \\
        StyleGAN-XL \citeyearpar{sauer2022stylegan} & 166M                   & --                & 1571                   & \; \; $1$
                                                    & 14.39                  & 2.20              & 15.57                  & \ \ 2.39           \\
        Drift-L
                                                    & 465M                   & \; CFG$^\ast$     & 87                     & \; \; $1$
                                                    & 10.51                  & 1.43              & \; --                  & --                 \\
        pMF-B \citeyearpar{lu2026one}
                                                    & 118M                   & \; CFG$^\ast$     & 66                     & \; \; $1$
                                                    & 13.83                  & 3.15              & 14.26                  & \ \ 3.63           \\
        pMF-L
                                                    & 410M                   & \; CFG$^\ast$     & 117                    & \; \; $1$
                                                    & \; 9.09                & 2.72              & \; 9.96                & \ \ 2.75           \\
        pMF-H
                                                    & 935M                   & \; CFG$^\ast$     & 271                    & \; \; $1$
                                                    & \; 6.87                & 2.29              & \; 8.24                & \ \ 2.48           \\
        \cmidrule(lr){2-9}
                                                    &                        & --                & 22                     & \; \; $1$
                                                    & 14.53                  & 5.20              & 15.17                  & \ \ 5.88           \\
                                                    &                        & --                & 22                     & \; \; $4$
                                                    & \; 8.58                & 4.89              & \; 9.35                & \ \ 6.04           \\
        \multirow{-3}{*}{\textbf{Sphere2-B}}
                                                    & \multirow{-3}{*}{161M} & CFG               & 22                     & \; \; $4 \times 2$
                                                    & \; 7.83                & 7.00              & \; 8.61                & \ \ 7.94           \\
        \arrayrulecolor{gray4}\cmidrule(lr){1-9}\arrayrulecolor{black}
                                                    &                        & --                & 115                    & \; \; $1$
                                                    & 11.96                  & 8.59              & 13.52                  & 10.46              \\
                                                    &                        & --                & 115                    & \; \; $4$
                                                    & \; 5.77                & 4.19              & \; 6.35                & \ \ 5.83           \\
        \multirow{-3}{*}{\textbf{Sphere2-L}}
                                                    & \multirow{-3}{*}{487M} & CFG               & 115                    & \; \; $4 \times 2$
                                                    & \; 5.46                & 5.38              & \; 6.27                & \ \ 7.10           \\
        \midrule
                                                    &                        & --                & 22                     & \; \; $1$
                                                    & 12.76                  & 3.43              & 13.91                  & \ \ 3.85           \\
                                                    &                        & --                & 22                     & \; \; $4$
                                                    & \; 7.95                & 3.07              & \; 8.49                & \ \ 2.92           \\
        \multirow{-3}{*}{\textbf{Sphere2-B}$^\dagger$}
                                                    & \multirow{-3}{*}{161M} & CFG               & 22                     & \; \; $4 \times 2$
                                                    & \; 6.91                & 3.90              & \; 7.82                & \ \ 4.47           \\
        \arrayrulecolor{gray4}\cmidrule(lr){1-9}\arrayrulecolor{black}
                                                    &                        & --                & 115                    & \; \; $1$
                                                    & 13.97                  & 5.14              & 11.65                  & \ \ 3.69           \\
                                                    &                        & --                & 115                    & \; \; $4$
                                                    & \; 5.14                & 2.44              & \; 6.60                & \ \ 3.02           \\
        \multirow{-3}{*}{\textbf{Sphere2-L}$^\dagger$}
                                                    & \multirow{-3}{*}{487M} & CFG               & 115                    & \; \; $4 \times 2$
                                                    & \; \second{4.92}       & 2.86              & \; \third{5.99}        & \ \ 3.53           \\
        \bottomrule
    \end{tabular}
\end{table}

On ImageNet, \cref{tab:fnl_imagenet} compares Sphere Encoder 2 with latent-space and pixel-space generators.
Since Sphere Encoder 2 generates in pixel space, we mainly focus on pixel-space baselines: PixNerd \citep{wang2026pixnerd}, JiT \citep{li2026back}, Drift \citep{deng2026generative}, and pMF \citep{lu2026one}, covering both one-step and many-step generation.

\begin{figure}[h]
   \centering
   \includegraphics[width=1.0\linewidth]{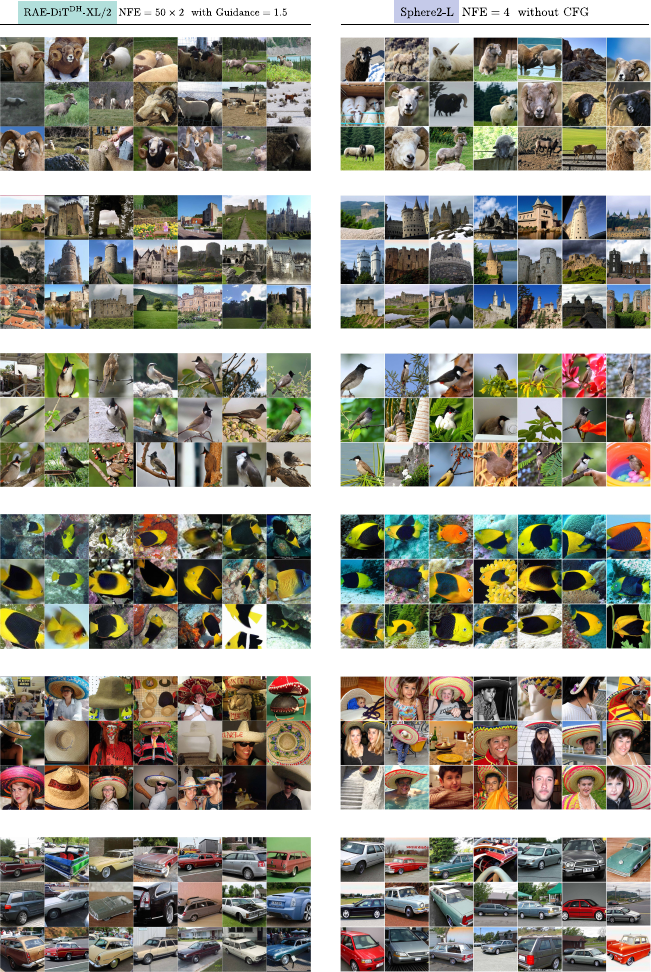}
   \caption{
      \textbf{Uncurated samples on ImageNet} $512 \times 512$.
      Left: RAE-DiT$^{\text{DH}}$-XL/2 with $\text{NFE} = 50 \times 2$ and $\text{Guidance} = 1.5$.
      Right: Sphere2-L with $\text{NFE} = 4$ and no CFG.
      Best viewed zoomed in.
   }
   \label{fig:sphere2_vs_rae}
\end{figure}

\begin{figure}[h]
   \centering
   \includegraphics[width=1.0\linewidth]{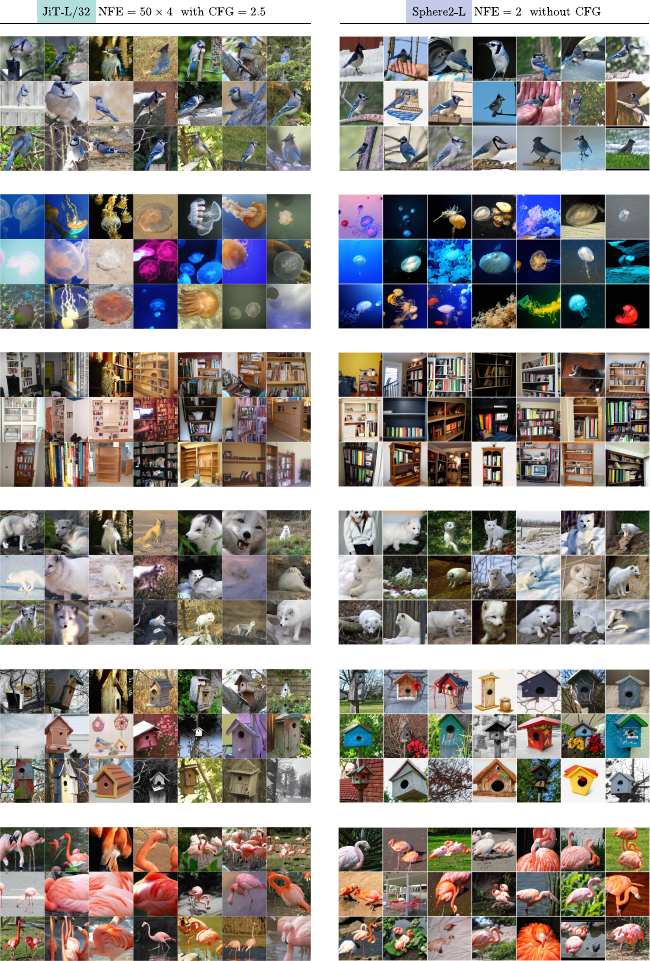}
   \caption{
      \textbf{Uncurated samples on ImageNet} $512 \times 512$.
      Left: JiT-L/32 with $\text{NFE} = 50 \times 4$ and $\text{CFG} = 2.5$.
      Right: Sphere2-L with $\text{NFE} = 2$ and no CFG.
      Best viewed zoomed in.
   }
   \label{fig:sphere2_vs_jit}
\end{figure}

\begin{figure}[h]
   \centering
   \includegraphics[width=1.0\linewidth]{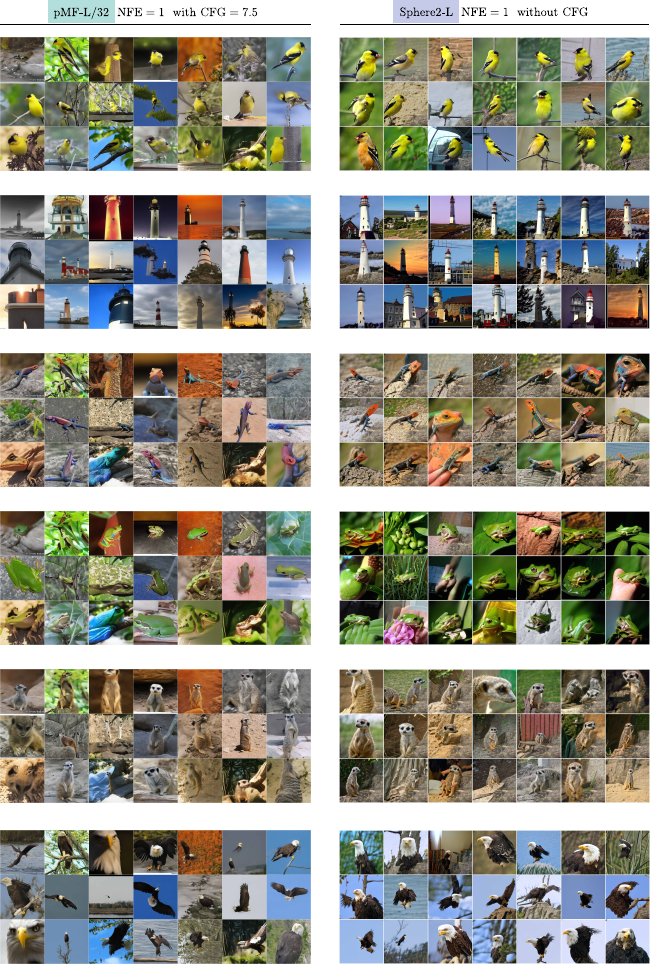}
   \caption{
      \textbf{Uncurated samples on ImageNet} $512 \times 512$.
      Left: pMF-L/32 with $\text{NFE} = 1$ and $\text{CFG} = 7.5$.
      Right: Sphere2-L with $\text{NFE} = 1$ and no CFG.
      Best viewed zoomed in.
   }
   \label{fig:sphere2_vs_pmf}
\end{figure}

At $256 \times 256$, Sphere2-B without CFG at 1 NFE is on par with JiT-B (14.53 vs. 15.78) and pMF-B (14.53 vs. 13.83), and at 4 NFE it outperforms JiT-L (8.58 vs. 10.73).
Sphere2-L at 4 NFE reaches 5.77, better than JiT-H (7.66) and pMF-H (6.87), and among pixel-space models is behind only many-step PixelDiT (4.46) and PixNerd (5.01), which use $66\times$ and $57\times$ more sampling compute, respectively.
The same ordering holds at $512 \times 512$.
\cref{fig:sphere2_vs_rae,fig:sphere2_vs_jit,fig:sphere2_vs_pmf} show uncurated $512 \times 512$ samples of Sphere2-L at 4, 2 and 1 steps without CFG, side by side with RAE-DiT$^\text{DH}$-XL/2 ($50 \times 2$ NFE), JiT-L ($50 \times 4$ NFE) and pMF-L (1 NFE), respectively.

While Sphere2 leads on FDr$^{6}$, its gFID is higher than that of other models.
We attribute the gap to the feature space rather than the loss: $\cL_\mathrm{score}$ matches real and fake distributions only in ConvNeXt V2-N features, and the match carries over to the other five modern spaces of FDr$^6$ (\cref{tab:fdr6_models}) but not to the old Inception space.
We investigate this in \cref{appsec:distribution_shift} and train a second group of Sphere2 models with an FD-lite loss $\cL_\mathrm{FD\text{-}lite}$ that mitigates this problem, marked by $^\dagger$ in \cref{tab:fnl_imagenet}; Sphere2 models without the mark are trained without it.
gFID drops for every model, and FDr$^{6}$ improves in most settings, almost entirely through its FID term; the other five feature spaces do not move much.

\begin{figure}[h]
   \centering
   \includegraphics[width=1.0\linewidth]{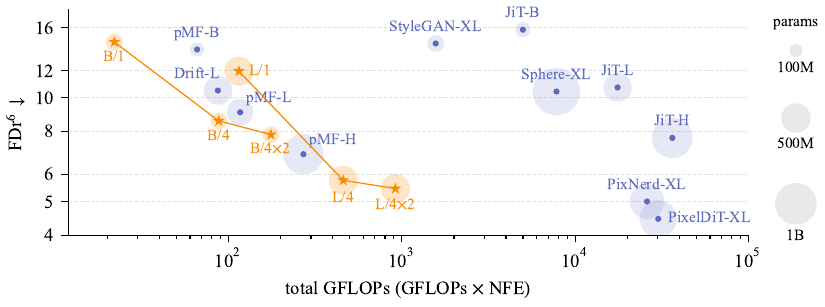}
   \caption{
      \textbf{Generation quality vs. sampling compute} on ImageNet $256 \times 256$ among pixel-space models.
      Circles are pixel-space \base{baselines} and stars are \ours{Sphere2} at 1, 4 (without CFG) and $4 \times 2$ NFE (with CFG), joined per model.
      The halo area around each marker is proportional to parameters.
   }
   \label{fig:fnl_imagenet_flops}
\end{figure}

\cref{fig:fnl_imagenet_flops} plots generation quality against sampling compute for the pixel-space models.
Without CFG, Sphere2-B sits below Drift-L and pMF-B/L at almost the same compute, and below JiT-B/L at far less compute, \ie, better FDr$^{6}$ for equal or fewer GFLOPs.
Sphere2-L without CFG reaches a lower FDr$^{6}$ than pMF-H at comparable compute, and than JiT-H at $79\times$ less.
\cref{tab:abl_speed} compares the sampling speed of the large models at $512 \times 512$ on a single H100, and adds RAE-DiT$^\text{DH}$-XL/2, the latent diffusion model with the best FDr$^{6}$ at this resolution in \cref{tab:fnl_imagenet}.
For sampling 50K images at batch size 16, Sphere2-L with CFG is $13\times$ faster than JiT-L and $66\times$ faster than RAE-DiT$^\text{DH}$-XL/2.

\begin{table}[h]
      \centering
      \caption{
            \textbf{Sampling speed} at $512 \times 512$ on a single NVIDIA H100.
            Timings are end-to-end, averaged over runs after warmup (std below $1\%$), at batch size 1 and 16.
            For Sphere2, ``loop $\times K$'' denotes $K$ sampling steps of \cref{alg:forward}.
            NFE counts decoder forwards; ``$+\,m$ enc.'' adds the encoder forwards used by spherical CFG, and ``$+\,49$ guide'' the guiding-model forwards of AutoGuidance.
            Sphere2 and JiT run in bf16, pMF and RAE-DiT in fp32, following their official implementations.
      }
      \label{tab:abl_speed}
      \small
      \setlength{\tabcolsep}{4.2pt}
      \begin{tabular}{l l c r r<{\hspace*{0.5em}} l r<{\hspace*{0.6em}} r<{\hspace*{2.0em}}}
            \toprule
            Model & Sampler            & Guid.          & bsz & \multicolumn{1}{c}{img/s $\uparrow$} & NFE           & \multicolumn{1}{c}{ms/NFE} & \multicolumn{1}{c}{50K imgs (min) $\downarrow$} \\
            \midrule
            \multirow{6}{*}{Sphere2-L}
                  & loop $\times$1     & --             & 1   & 53.4                                 & 1             & 18.7                       & 16                                              \\
                  & loop $\times$2     & CFG            & 1   & 14.5                                 & 4 + 2 enc.    & 17.3                       & 58                                              \\
                  & loop $\times$4     & CFG            & 1   & 6.1                                  & 8 + 6 enc.    & 20.7                       & 137                                             \\
                  & loop $\times$1     & --             & 16  & 424.8                                & 1             & 2.4                        & 2                                               \\
                  & loop $\times$2     & CFG            & 16  & 145.4                                & 4 + 2 enc.    & 1.7                        & 6                                               \\
                  & loop $\times$4     & CFG            & 16  & 62.8                                 & 8 + 6 enc.    & 2.0                        & 14                                              \\
            \midrule
            \multirow{2}{*}{JiT-L/32}
                  & Heun $\times$50    & CFG            & 1   & 0.5                                  & 198           & 10.3                       & 1666                                            \\
                  & Heun $\times$50    & CFG            & 16  & 4.5                                  & 198           & 1.1                        & 185                                             \\
            \midrule
            \multirow{2}{*}{pMF-L/32}
                  & MeanFlow $\times$1 & \ \ CFG$^\ast$ & 1   & 42.9                                 & 1             & 23.3                       & 20                                              \\
                  & MeanFlow $\times$1 & \ \ CFG$^\ast$ & 16  & 157.7                                & 1             & 6.3                        & 6                                               \\
            \midrule
            \multirow{2}{*}{RAE-DiT$^{\text{DH}}$-XL/2}
                  & Euler $\times$49   & \ \ AG         & 1   & 0.6                                  & 49 + 49 guide & 30.8                       & 1389                                            \\
                  & Euler $\times$49   & \ \ AG         & 16  & 0.9                                  & 49 + 49 guide & 22.1                       & 926                                             \\
            \bottomrule
      \end{tabular}
\end{table}

\section{Conclusion}
Sphere Encoders are an emerging paradigm with progress from several teams \citep{do2026efficient,man2026sp}, generating images with few steps and low compute.
Sphere Encoder 2 takes this paradigm a step further, matching the image quality of more expensive diffusion models while keeping that efficiency.
It closes the gap to the equator with an explicit rotation, and splits the rotation arc into reconstruction and generation regimes on the same sphere.
The generation regime is trained with a simple objective: semantic alignment and latent consistency keep generated images semantically faithful to the input but structurally different, and a lightweight score matching loss sharpens them without adversarial training.
Several other families also targeted one-step generation from scratch, including consistency models, MeanFlow and its variants, drifting models, and GANs.
Sphere Encoder 2 adds a complementary path to this landscape by bringing the autoencoder back as a generator in its own right and could be a viable candidate for one/few-step generation.

\begin{figure}[h]
   \centering
   \includegraphics[width=1.0\linewidth]{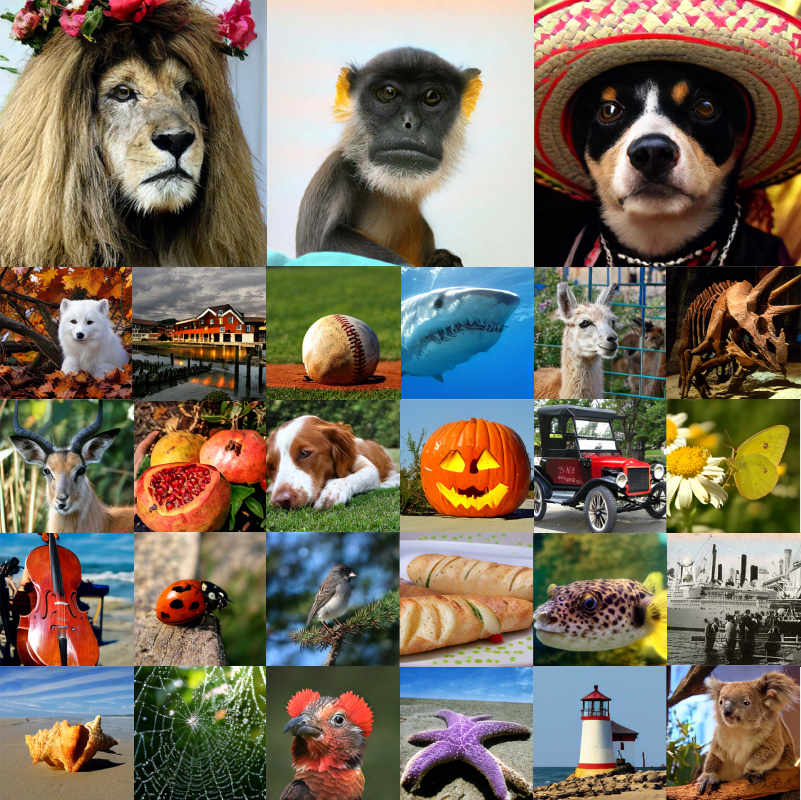}
   \caption{
      \textbf{Curated 2-step generation} without CFG on ImageNet $512 \times 512$ by Sphere Encoder 2.
   }
   \label{fig:imagenet_512px_2step}
\end{figure}

\vspace{-0.5em}
\section*{Acknowledgments}
This work was supported by the NSF TRAILS Institute (2229885), and Coefficient Giving. This material is based upon work supported by the U.S. Department of Energy, Office of Science, Office of Advanced Scientific Computing Research under the Advancements in AI for Science program. This research used resources from the Argonne Leadership Computing Facility, a U.S. DOE Office of Science user facility at Argonne National Laboratory, which is supported by the Office of Science of the U.S. DOE under Contract No. DE-AC02-06CH11357.

Prepared by LLNL under Contract DE-AC52-07NA27344 and supported by the LLNL-LDRD Program under Project No. 24-ERD-010 (LLNL-PROC-2025006). This manuscript has been authored by Lawrence Livermore National Security, LLC under Contract No. DE-AC52-07NA27344 with the U.S. Department of Energy. The United States Government retains, and the publisher, by accepting the article for publication, acknowledges that the United States Government retains a non-exclusive, paid-up, irrevocable, world-wide license to publish or reproduce the published form of this manuscript, or allow others to do so, for United States Government purposes.

\clearpage

\clearpage
\appendix
\renewcommand{\thefigure}{A\arabic{figure}}
\renewcommand{\thetable}{A\arabic{table}}
\renewcommand{\theequation}{A\arabic{equation}}

\section*{Appendix}

\section{Interpolation}

\begin{figure}[h]
    \centering
    \includegraphics[width=1.0\linewidth]{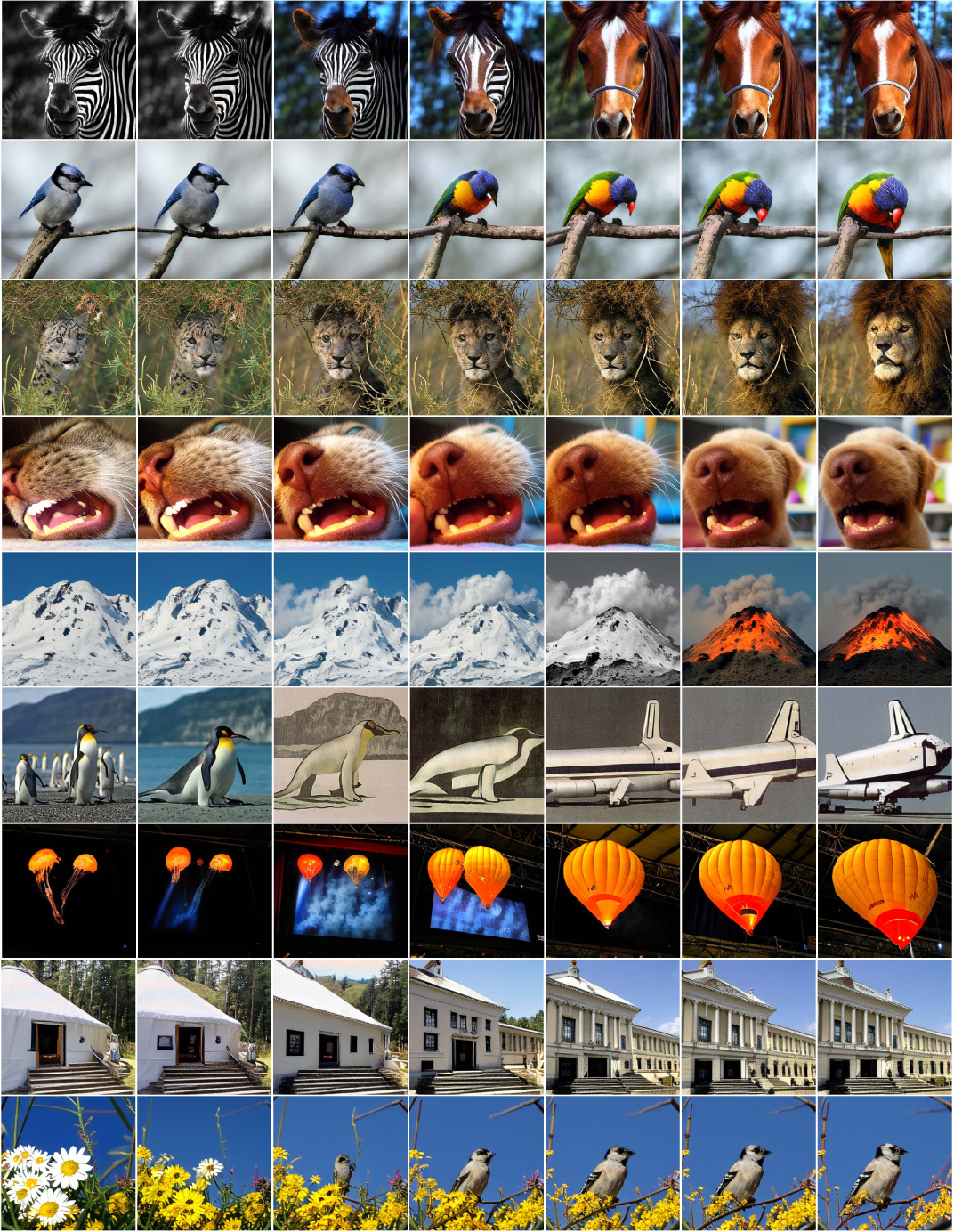}
    \caption{
        \textbf{Class interpolation on ImageNet} $512 \times 512$ by Sphere2-L with 4-step sampling and no CFG.
        Each row fixes the random point and slerps between two class embeddings, from the left class to the right one.
        Rows are ordered from similar pairs at the top to dramatic pairs at the bottom.
    }
    \label{fig:imagenet_512px_interp}
\end{figure}

\cref{fig:imagenet_512px_interp} shows class interpolation: $\cD(\cF(\be), \y, \alpha)$.
The random point $\cF(\be)$ is fixed in \cref{eq:generation_conditional} and we interpolate points between two class embeddings $\y_a,\y_b$ by spherical linear interpolation (slerp).

\begin{figure}[h]
    \centering
    \includegraphics[width=1.0\linewidth]{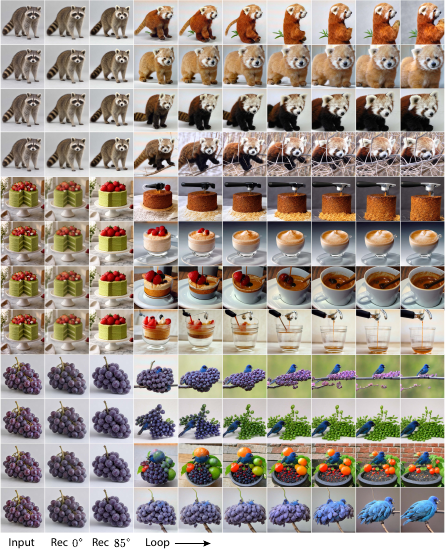}
    \caption{
        \textbf{Interpolation from non-ImageNet images} by Sphere2-L trained on ImageNet $512 \times 512$.
        The first column shows input images \emph{outside} the ImageNet classes: a raccoon, a matcha cake, and grapes.
        The second and third columns decode the clean latent of each input conditioned on $\alpha = 0^\circ$ and $\alpha = 85^\circ$, respectively.
        The remaining columns unroll the refinement loop of \cref{alg:forward}, which carries the input toward a target class: raccoon to red panda (class 387), matcha cake to espresso (class 967), and grapes to indigo bunting (class 14).
    }
    \label{fig:imagenet_512px_interp_by_recon}
\end{figure}

\cref{fig:imagenet_512px_interp_by_recon} interpolates from images \emph{outside} ImageNet.
The input $\X$ is looped under an ImageNet class starting from $\cD(\cE(\X), \y, \alpha)$, then \cref{alg:forward} carries it toward target class $\y$.
At $\alpha = 0^\circ$, in the reconstruction regime, the decoder reconstructs the input from its clean latent.
The reconstruction shows the waffle-like texture seen in \cref{fig:abl_cutoff_vis_flowers}, which is expected since this regime omits adversarial sharpening.
At $\alpha = 85^\circ$, in the generation regime, the output is sharper and drifts toward the ImageNet manifold.
Each further refinement step moves it closer to the target class.

\section{Ablation Studies}
\label{appsec:ablations}

\subsection{Distribution Shift}
\label{appsec:distribution_shift}

$\cL_\mathrm{score}$ matches the fake and real distributions in one feature space, ConvNeXt V2-N.
As \cref{sec:experiments} notes, the match carries over to the five modern spaces of FDr$^6$ but not to the old Inception space of gFID, where the statistics of generated images drift away from those of real images.
This is a mirror image of the metric paradox in FDr$^6$ \citep{yang2026representation}: most generators beat gFID but not the eye, while ours satisfies the modern feature spaces but not gFID.
Ideally, we could learn a feature space specific to ImageNet as drifting models do \citep{deng2026generative}, or add score models in the Inception space, but both are expensive and cumbersome.
We seek a simple and cheap alternative.

To pin the statistics in place, we borrow an idea from FD loss \citep{yang2026representation} and add a Fréchet distance term on frozen features as a regularizer:
\begin{align}
    \cL_\mathrm{FD\text{-}lite}
    = \sum_i \frac{
        \ \mathrm{FD}_{\phi_i}(\hat{\X}, \X) \phantom{2}
    }{
        \phantom{1} \mathrm{sg}\big(\mathrm{FD}_{\phi_i}(\hat{\X}, \X)\big) + \epsilon
    },
\end{align}
where $\epsilon=10^{-6}$ and $\mathrm{FD}_{\phi_i}$ is the Fréchet distance \citep{heusel2017gans} between real and generated features from the $i^{\mathrm{th}}$ frozen extractor $\phi_i$.
Each term is normalized by its own detached value, so every extractor contributes a gradient of unit scale.
The mean and covariance of real and fake features are tracked with exponential moving averages (EMA) online, and $\mathrm{FD}_{\phi_i}$ is computed between them.
We use three small extractors: Inception, our encoder $\cE$ (DINOv3-small+, 29M), and our score feature extractor $\phi$ (ConvNeXt V2-N, 16M).
The latter two are already run for $\cL_\mathrm{lat\text{-}con}$ and $\cL_\mathrm{score}$, so $\cL_\mathrm{FD\text{-}lite}$ adds one Inception forward and almost no compute.

\begin{table}[h]
    \centering
    \caption{
        \textbf{Effect of $\cL_\mathrm{FD\text{-}lite}$ per feature space} on ImageNet $256 \times 256$.
        FDr$^{6}$ at $K$ sampling steps, trained with and without $\cL_\mathrm{FD\text{-}lite}$, broken down into its six feature spaces.
    }
    \label{tab:abl_distribution_shift_per_space}
    \small
    \setlength{\tabcolsep}{6.4pt}
    \begin{tabular}{c c c c cccccc}
        \toprule
        \multirow{2}{*}[-0.5ex]{Steps $K$}                     &
        \multirow{2}{*}[-0.5ex]{$\cL_\mathrm{FD\text{-}lite}$} &
        \multirow{2}{*}[-0.5ex]{gFID $\downarrow$}             &
        \multirow{2}{*}[-0.5ex]{FDr$^6$ $\downarrow$}          &
        \multicolumn{6}{c}{FDr$^{6}$ $\downarrow$ per feature space}                                                                              \\
        \cmidrule(l){5-10}
                                                               &
                                                               &            &         & Incep. & ConvNeXt & DINOv2  & MAE     & SigLIP  & CLIP    \\
        \midrule
        \rowcolor{gray1}
        \multicolumn{10}{l}{\textit{Sphere2-B}}                                                                                                   \\
                                                               & -
                                                               & 5.20       & 14.53   & 3.49   & 2.99     & 13.18   & 15.08   & 28.01   & 24.42   \\
        \multirow{-2}{*}{1}                                    & \checkmark
                                                               & 3.43       & 12.76   & 2.30   & 2.59     & 11.55   & 13.64   & 24.41   & 22.03   \\
        \cmidrule(lr){1-10}
                                                               & -
                                                               & 4.72       & \; 9.48 & 3.17   & 2.67     & \; 8.83 & 10.80   & 16.09   & 15.35   \\
        \multirow{-2}{*}{2}                                    & \checkmark
                                                               & 2.69       & \; 8.43 & 1.81   & 2.36     & \; 7.43 & \; 9.62 & 14.83   & 14.51   \\
        \cmidrule(lr){1-10}
                                                               & -
                                                               & 4.89       & \; 8.58 & 3.28   & 3.35     & \; 7.87 & 10.05   & 13.51   & 13.41   \\
        \multirow{-2}{*}{4}                                    & \checkmark
                                                               & 3.07       & \; 7.95 & 2.06   & 3.39     & \; 6.87 & \; 8.82 & 13.38   & 13.19   \\
        \midrule
        \rowcolor{purple0}
        \multicolumn{10}{l}{\textit{Sphere2-L}}                                                                                                   \\
                                                               & -
                                                               & 8.59       & 11.96   & 5.76   & 2.49     & \; 9.18 & 13.54   & 21.29   & 19.48   \\
        \multirow{-2}{*}{1}                                    & \checkmark
                                                               & 5.14       & 13.97   & 3.45   & 3.11     & \; 9.01 & 14.65   & 29.71   & 23.89   \\
        \cmidrule(lr){1-10}
                                                               & -
                                                               & 5.77       & \; 6.42 & 3.88   & 1.78     & \; 5.43 & \; 7.39 & \; 9.62 & 10.41   \\
        \multirow{-2}{*}{2}                                    & \checkmark
                                                               & 3.13       & \; 5.92 & 2.10   & 1.70     & \; 4.68 & \; 6.97 & \; 9.75 & 10.31   \\
        \cmidrule(lr){1-10}
                                                               & -
                                                               & 4.19       & \; 5.77 & 2.81   & 2.69     & \; 5.00 & \; 6.56 & \; 8.50 & \; 9.05 \\
        \multirow{-2}{*}{4}                                    & \checkmark
                                                               & 2.44       & \; 5.14 & 1.64   & 2.05     & \; 4.15 & \; 6.01 & \; 8.16 & \; 8.86 \\
        \bottomrule
    \end{tabular}
\end{table}

\textbf{Where the drift is.}
We train Sphere2-B and Sphere2-L on ImageNet $256 \times 256$ with and without $\cL_\mathrm{FD\text{-}lite}$ for the same epochs, and report FDr$^{6}$ in each feature space in \cref{tab:abl_distribution_shift_per_space}.
The Inception space is where the regularizer acts: its FDr drops by $34$--$46\%$ for both models at every step count (Incep. column in \cref{tab:abl_distribution_shift_per_space}), which is the gFID improvement in \cref{tab:fnl_imagenet}.
The other five modern spaces move far less, mostly by single digits, confirming that the drift is specific to the decade-old Inception space rather than a general mismatch.

\begin{figure}[h]
    \centering
    \includegraphics[width=1.0\linewidth]{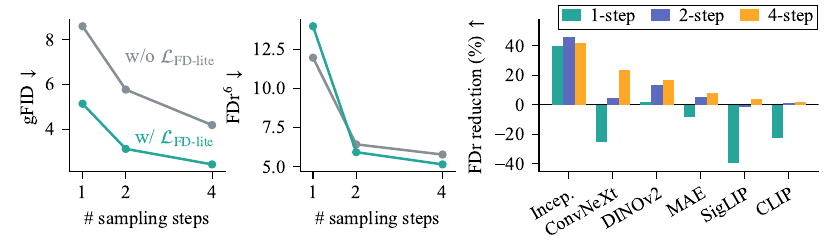}
    \caption{
        \textbf{Effect of $\cL_\mathrm{FD\text{-}lite}$} for Sphere2-L on ImageNet $256 \times 256$.
        Left and middle: FID and FDr$^{6}$ with and without the loss at different sampling steps.
        Right: relative FDr reduction in each of the six feature spaces.
    }
    \label{fig:abl_distribution_shift_res}
\end{figure}

\textbf{Model size matters.}
Sphere2-B benefits in every space at every step count.
Sphere2-L benefits at 2 and 4 steps, but at 1 step its overall FDr$^{6}$ gets worse by $17\%$: the Inception term improves as before, but ConvNeXt, MAE, SigLIP and CLIP all degrade (\cref{fig:abl_distribution_shift_res}, right).
We read this as a capacity effect.
The smaller decoder has slack to satisfy the extra constraint, while the larger decoder, which already matches the modern spaces closely without the regularizer, trades some one-step fidelity in those spaces for the Inception match.
Multi-step sampling recovers the loss, since each refinement step re-encodes the image and pulls it back toward the real manifold.

\begin{figure}[h]
    \centering
    \includegraphics[width=1.0\linewidth]{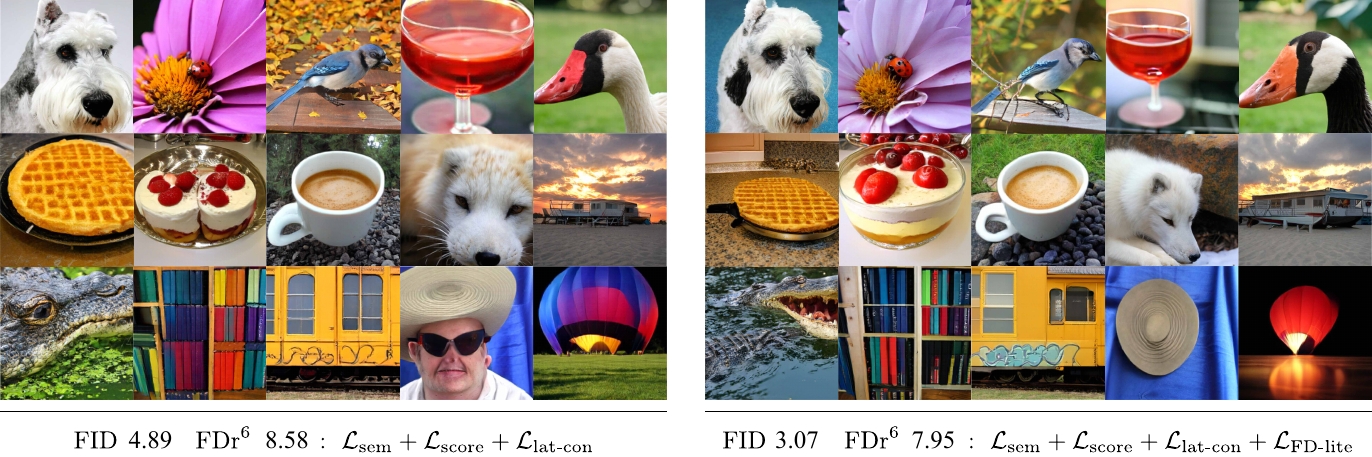}
    \caption{
        \textbf{Distribution shift is invisible but measurable.}
        4-step samples on ImageNet $256 \times 256$ from Sphere2-B trained with and without $\cL_\mathrm{FD\text{-}lite}$ are visually indistinguishable, yet the loss aligns their feature statistics with those of real data in the Inception space where gFID is measured.
        More uncurated samples for side by side comparison are in \cref{fig:fig_w_vs_wo_fd_loss_1step} for 1 step and \cref{fig:fig_w_vs_wo_fd_loss_4step} for 4 steps.
    }
    \label{fig:abl_distribution_shift_vis}
\end{figure}

\textbf{Invisible to the eye.}
\cref{fig:abl_distribution_shift_vis} compares generations with and without $\cL_\mathrm{FD\text{-}lite}$ from Sphere2-B, the model it affects more.
The two sets are visually indistinguishable
\footnote{
    We release Sphere2 checkpoints trained with and without $\cL_\mathrm{FD\text{-}lite}$ in our repo for readers' own FDr\textsuperscript{\scalebox{1.3}{\twemoji{eyes}}}.
}
, yet the loss aligns their feature statistics with those of real data in the space where gFID lives.
A good metric for image generation is still an open problem, and we hope our results on ImageNet encourage more work on it.
If a model is judged by human preference or prompt-following benchmarks rather than gFID, as text-to-image models commonly are, we believe $\cL_\mathrm{FD\text{-}lite}$ is unnecessary and can be dropped.

\textbf{Relation to FD loss.}
We call ours $\cL_\mathrm{FD\text{-}lite}$ because it differs from the original FD loss \citep{yang2026representation} in purpose, and as a result in cost.
The original is a post-training objective: it is the only loss applied to a pretrained multi-step generator, and its job is to repurpose the generator for one-step generation, which visibly and dramatically improves one-step quality.
Ours is a regularizer.
It runs alongside our other three losses throughout training with a small weight of $0.1$, and only keeps the statistics of already plausible samples from drifting particularly in the Inception space.
Being a regularizer rather than the main objective, it does not need the heavy extractors of the original, Inception, SigLIP-so400M and MAE-Large, and reuses the two we already run in our loss functions.

\begin{figure}[h]
    \centering
    \includegraphics[width=1.0\linewidth]{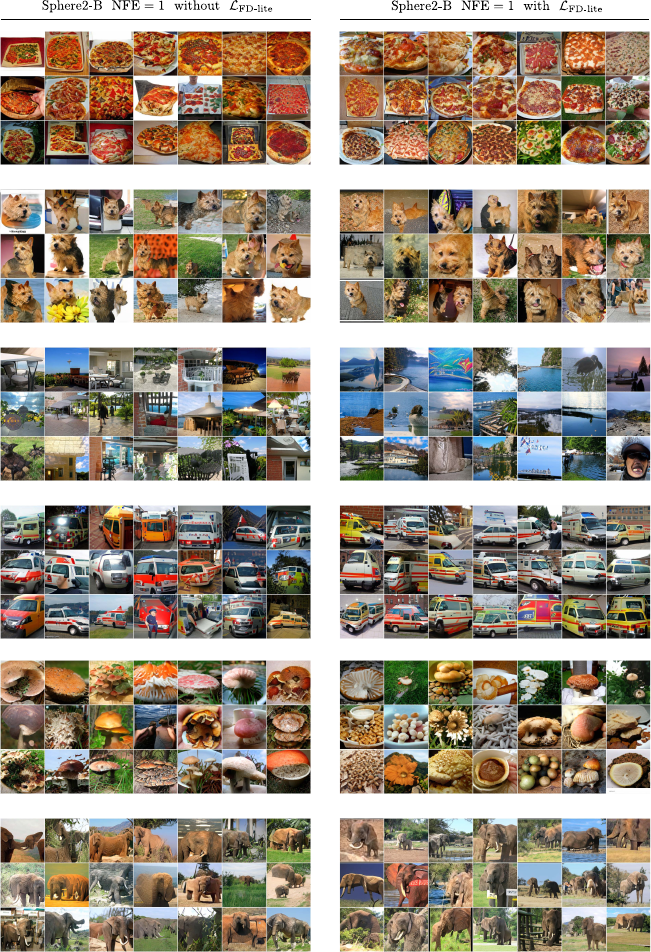}
    \caption{
        \textbf{Uncurated samples on ImageNet} $256 \times 256$.
        Left: Sphere2-B without $\cL_\mathrm{FD\text{-}lite}$ and $\mathrm{NFE} = 1$.
        Right: Sphere2-B with $\cL_\mathrm{FD\text{-}lite}$ and $\mathrm{NFE} = 1$.
    }
    \label{fig:fig_w_vs_wo_fd_loss_1step}
\end{figure}

\begin{figure}[h]
    \centering
    \includegraphics[width=1.0\linewidth]{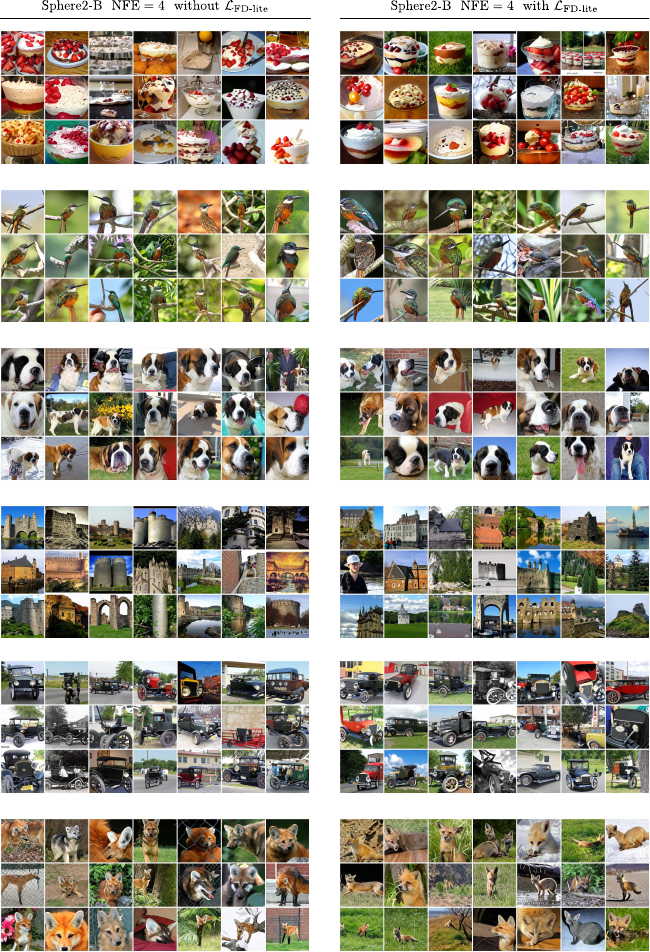}
    \caption{
        \textbf{Uncurated samples on ImageNet} $256 \times 256$.
        Left: Sphere2-B without $\cL_\mathrm{FD\text{-}lite}$ and $\mathrm{NFE} = 4$.
        Right: Sphere2-B with $\cL_\mathrm{FD\text{-}lite}$ and $\mathrm{NFE} = 4$.
    }
    \label{fig:fig_w_vs_wo_fd_loss_4step}
\end{figure}

\clearpage
\subsection{Spherical CFG}
\label{appsec:cfg}

Classic CFG \citep{ho2022classifier} linearly extrapolates from the unconditional prediction to the conditional one, which would push a latent off the sphere.
We instead guide on the sphere with a rotation.
From the same noise $\cF(\be)$, we follow two parallel arcs of \cref{alg:forward}: a conditional arc with class $\y$ and an unconditional arc with the null class $\y_{\mathrm{null}}$.
Following common practice, the null class is learned during training by dropping the class embedding to $\y_{\mathrm{null}}$ with probability $0.1$.
At each refinement step, we encode both generated images into latents $\z_c$ and $\z_u$, and rotate $\z_c$ away from $\z_u$ along their great circle by a fixed guidance angle $\gamma$:
\begin{align}
    \label{eq:spherical_cfg}
    \z_\mathrm{cfg}
    = \frac{\sin(\theta + \gamma)}{\sin\theta} \, \z_c - \frac{\sin\gamma}{\sin\theta} \, \z_u,
    \qquad \theta = \angle(\z_u, \z_c).
\end{align}
The guided latent $\z_\mathrm{cfg}$ stays on the sphere at angle $\theta + \gamma$ from $\z_u$, and replaces $\z_c$ in the rotation toward the noise in \cref{eq:rotation}.\footnote{
    When $\z_c$ and $\z_u$ are parallel or antiparallel ($\sin\theta = 0$), no rotation plane is defined, and we keep $\z_\mathrm{cfg} = \z_c$.
    This never occurs in practice, since the two arcs decode different images in the high-dimensional latent space.
}
Unlike a guidance scale, $\gamma$ fixes the step on the sphere regardless of how far apart the two arcs are, and $\gamma = 0$ recovers \cref{alg:forward}.
The unconditional arc is never guided and continues from its own latent $\z_u$; if it were decoded from $\z_\mathrm{cfg}$, the two arcs would merge after one step and leave nothing to guide against.
Since guidance acts on the refinement steps, it requires $K \geq 2$, and one-step generation stays unguided.
\cref{alg:cfg} summarizes the sampling procedure.

\noindent
\begin{algorithm}[H]
    \caption{
        Sampling with spherical CFG. \texttt{rotate\_away} is \cref{eq:spherical_cfg} and \texttt{gamma} is the guidance angle.
        \texttt{gamma} $= 0^\circ$ disables guidance and recovers \cref{alg:forward}.
    }
    \label{alg:cfg}
    \lstset{
        backgroundcolor=\color{white},
        basicstyle=\fontsize{8.0pt}{7.4pt}\ttfamily\selectfont,
        columns=fullflexible,
        keepspaces=true,
        breaklines=true,
        captionpos=b,
        commentstyle=\fontsize{7.5pt}{7.5pt}\color{purple4},
        keywordstyle=\fontsize{7.5pt}{7.5pt},
        aboveskip=0.4em,
        belowskip=0.1em,
    }
    \begin{lstlisting}[language=python]
e = F(Normal(0, 1).sample([d])) # random point

# one-step generation
x   = D(e, y,      alpha) # conditional arc
x_u = D(e, y_null, alpha) # unconditional arc

# few-step refinement
for _ in range(K - 1):
    z_c, z_u = F(E(x)), F(E(x_u))
    z = rotate_away(z_c, z_u, gamma) # guided on the sphere
    
    x   = D(cos(alpha) * z   + sin(alpha) * e, y,      alpha)
    x_u = D(cos(alpha) * z_u + sin(alpha) * e, y_null, alpha)

return x
         \end{lstlisting}
\end{algorithm}

\begin{figure}[h]
    \centering
    \includegraphics[width=1.0\linewidth]{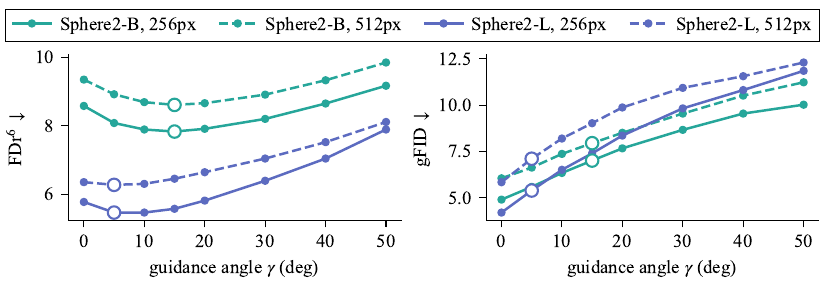}
    \caption{
        \textbf{Guidance angle $\gamma$} of spherical CFG at 4 sampling steps on ImageNet $256 \times 256$ and $512 \times 512$ in \cref{tab:abl_spherical_cfg}.
        The hollow marker on each curve is the FDr$^{6}$-optimal $\gamma$, which the CFG rows of \cref{tab:fnl_imagenet} use.
        Sampling angle $\alpha$ is $85.5^\circ$ for Sphere2-B and $84^\circ$ for Sphere2-L at both resolutions.
        See \cref{fig:abl_cfg_vis} for generated images along the $\gamma$ sweep.
    }
    \label{fig:abl_cfg_res}
\end{figure}

\begin{figure}[H]
    \vspace{-0.5em}
    \centering
    \includegraphics[width=1.0\linewidth]{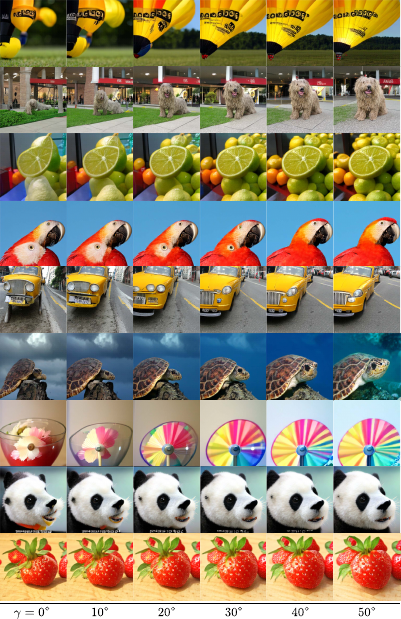}
    \caption{
        \textbf{Spherical CFG} for 2-step generation on ImageNet $256 \times 256$ with Sphere2-B.
        Each row is one random point decoded at increasing guidance angle $\gamma$.
        Guidance fixes the bad structures and strengthens the class semantics.
    }
    \label{fig:abl_cfg_vis}
\end{figure}

\begin{table}[h]
    \centering
    \caption{
        \textbf{Guidance angle $\gamma$} of spherical CFG on ImageNet $256 \times 256$ and $512 \times 512$.
        Sampling angle $\alpha$ is fixed per model, listed as $256 / 512$ when it differs between resolutions.
    }
    \label{tab:abl_spherical_cfg}
    \footnotesize
    \setlength{\tabcolsep}{3pt}
    \resizebox{\linewidth}{!}{%
        \begin{tabular}{l c c l cccccccc cccccccc}
            \toprule
            \multirow{2}{*}[-0.5ex]{Model}                                &
            \multirow{2}{*}[-0.5ex]{$\alpha$}                             &
            \multirow{2}{*}[-0.5ex]{Steps}                                &
            \multirow{2}{*}[-0.5ex]{Metric}                               &
            \multicolumn{8}{c}{Guidance angle $\gamma$, $256 \times 256$} &
            \multicolumn{8}{c}{Guidance angle $\gamma$, $512 \times 512$}                                                                                                                                                                                                                                                                                                                                                                 \\
            \cmidrule(lr){5-12} \cmidrule(lr){13-20}
                                                                          &                               &                    &                        & $0^\circ$        & $5^\circ$        & $10^\circ$       & $15^\circ$       & $20^\circ$       & $30^\circ$       & $40^\circ$ & $50^\circ$ & $0^\circ$        & $5^\circ$        & $10^\circ$       & $15^\circ$       & $20^\circ$ & $30^\circ$       & $40^\circ$ & $50^\circ$ \\
            \midrule
            \multirow{4}{*}{Sphere2-B}                                    & \multirow{4}{*}{$85.5^\circ$} & \multirow{2}{*}{2} & FDr$^{6}$ $\downarrow$ & \; 9.48          & \; 9.03          & \; 8.77          & \; 8.63          & \; \textbf{8.55} & \; \textbf{8.55} & \; 8.67    & \; 8.89    & 10.28            & \; 9.85          & \; 9.56          & \; 9.45          & \; 9.33    & \; \textbf{9.26} & \; 9.35    & \; 9.52    \\
                                                                          &                               &                    & gFID $\downarrow$      & \; \textbf{4.72} & \; 5.30          & \; 5.80          & \; 6.30          & \; 6.78          & \; 7.66          & \; 8.34    & \; 9.04    & \; \textbf{5.64} & \; 6.11          & \; 6.46          & \; 7.06          & \; 7.46    & \; 8.24          & \; 9.16    & \; 9.81    \\
            \cmidrule(lr){3-20}
                                                                          &                               & \multirow{2}{*}{4} & FDr$^{6}$ $\downarrow$ & \; 8.58          & \; 8.08          & \; 7.89          & \; \textbf{7.83} & \; 7.91          & \; 8.20          & \; 8.65    & \; 9.17    & \; 9.35          & \; 8.92          & \; 8.69          & \; \textbf{8.61} & \; 8.66    & \; 8.91          & \; 9.33    & \; 9.85    \\
                                                                          &                               &                    & gFID $\downarrow$      & \; \textbf{4.89} & \; 5.58          & \; 6.33          & \; 7.00          & \; 7.66          & \; 8.66          & \; 9.54    & 10.02      & \; \textbf{6.04} & \; 6.62          & \; 7.36          & \; 7.94          & \; 8.50    & \; 9.54          & 10.51      & 11.23      \\
            \midrule
            \multirow{4}{*}{Sphere2-L}                                    & \multirow{4}{*}{$84^\circ$}   & \multirow{2}{*}{2} & FDr$^{6}$ $\downarrow$ & \; 6.42          & \; 6.31          & \; \textbf{6.28} & \; 6.30          & \; 6.39          & \; 6.62          & \; 6.94    & \; 7.32    & \; 7.50          & \; 7.37          & \; \textbf{7.28} & \; 7.30          & \; 7.31    & \; 7.44          & \; 7.61    & \; 7.84    \\
                                                                          &                               &                    & gFID $\downarrow$      & \; \textbf{5.77} & \; 6.82          & \; 7.82          & \; 8.53          & \; 9.25          & 10.52            & 11.53      & 12.42      & \; \textbf{7.59} & \; 8.62          & \; 9.35          & 10.17            & 10.74      & 11.78            & 12.51      & 13.05      \\
            \cmidrule(lr){3-20}
                                                                          &                               & \multirow{2}{*}{4} & FDr$^{6}$ $\downarrow$ & \; 5.77          & \; \textbf{5.46} & \; \textbf{5.46} & \; 5.57          & \; 5.81          & \; 6.39          & \; 7.04    & \; 7.89    & \; 6.35          & \; \textbf{6.27} & \; 6.30          & \; 6.45          & \; 6.64    & \; 7.04          & \; 7.52    & \; 8.11    \\
                                                                          &                               &                    & gFID $\downarrow$      & \; \textbf{4.19} & \; 5.38          & \; 6.50          & \; 7.39          & \; 8.35          & \; 9.82          & 10.81      & 11.85      & \; \textbf{5.83} & \; 7.10          & \; 8.19          & \; 9.02          & \; 9.87    & 10.93            & 11.56      & 12.30      \\
            \bottomrule
        \end{tabular}%
    }
\end{table}

\subsection{Angle Distribution in Training}

\begin{figure}[h]
    \centering
    \includegraphics[width=1.0\linewidth]{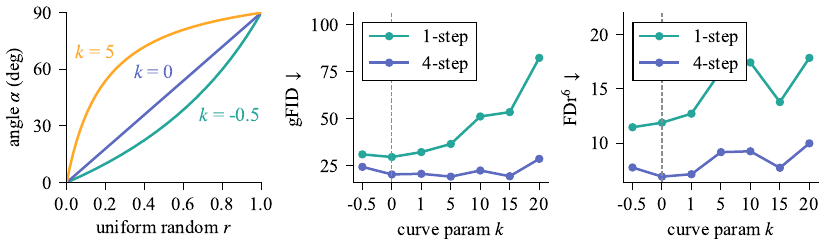}
    \caption{
        \textbf{Angle distribution in training.}
        Left: sampling curves of $\alpha$ for different $k$.
        Middle and right: generation quality of Sphere2-B on Oxford Flowers $256 \times 256$ for each $k$.
        Uniform sampling ($k = 0$) is the best choice.
    }
    \label{fig:abl_angle_curve}
\end{figure}

Diffusion models commonly concentrate training samples in the high-noise region, via a noise-level \citep{li2026back} or timestep shift \citep{esser2024scaling}.
Sphere Latent Encoder \citep{do2026efficient} likewise found that concentrating samples at high-angle noise improves generation.
One might expect Sphere Encoder 2 to benefit from more training at high angles as well, especially since its decoder is conditioned on $\alpha$.
To test this expectation, we train a Sphere2-B on Oxford Flowers $256 \times 256$ and control the sampling density of $\alpha$ during training with a single parameter by drawing it from a rational curve:
\begin{align}
    \alpha = 90^\circ \cdot \frac{(k + 1)\, r}{k\, r + 1},
    \quad r \sim \mathcal{U}[0, 1],
\end{align}
where $k > -1$ sets the shape of the curve (\cref{fig:abl_angle_curve} left).
Setting $k = 0$ gives uniform sampling.
A positive $k$ concentrates samples near the equator, and a negative $k$ near the pole.
\cref{fig:abl_angle_curve} middle and right tell a different story from common practice.

Uniform sampling ($k = 0$) gives the best 1-step gFID, and the best FDr$^{6}$ at every step count.
Concentrating samples near the equator ($k > 0$) buys only a marginal gain in multi-step gFID at $k = 5$.
We attribute this to the angle cutoff: each regime already receives a loss tailored to it, so the decoder no longer needs an uneven sampling density to prioritize the generation.
We therefore use uniform sampling by default, as the simplest and most effective choice.

\subsection{Maximum Training Angle}
\label{appsec:abl_max_angle}

\begin{figure}[h]
    \centering
    \includegraphics[width=1.0\linewidth]{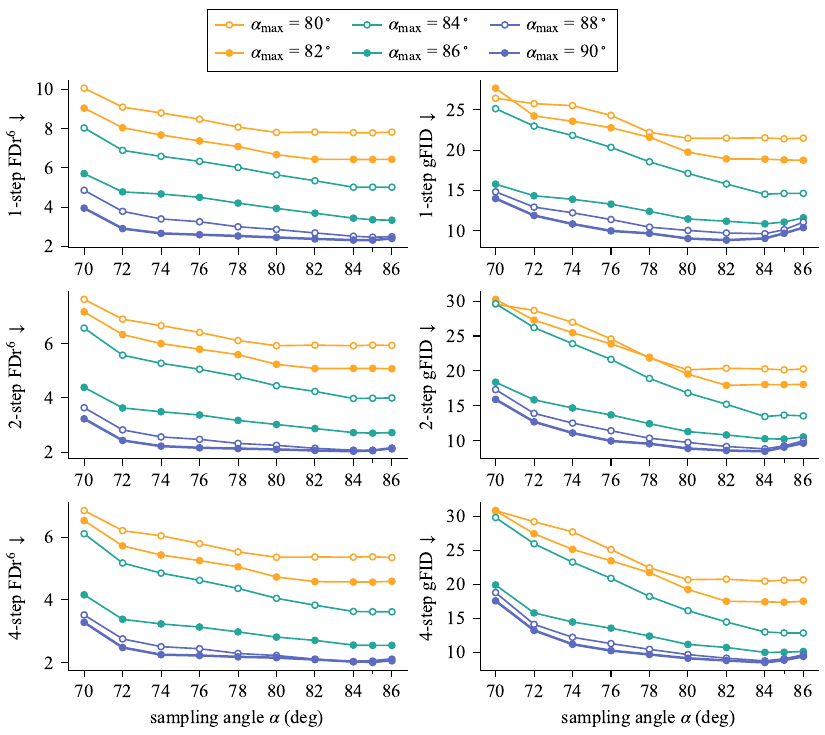}
    \caption{
        \textbf{Maximum training angle $\alpha_\mathrm{max}$} of Sphere2-B on Oxford Flowers $256 \times 256$.
        Left and right: FDr$^{6}$ and gFID along the sampling angle $\alpha$ at inference, one curve per $\alpha_\mathrm{max}$.
        Top to bottom: 1, 2 and 4 sampling steps.
    }
    \label{fig:abl_max_angle_res}
\end{figure}

In \cref{sec:sphere2}, we hypothesize that the maximum training angle $\alpha_\mathrm{max}$ in \cref{eq:rotation} should be exactly $90^\circ$ to reach the equator, where the random points of generation live.
\cref{fig:abl_max_angle_res} ablates $\alpha_\mathrm{max}$ with Sphere2-B on Oxford Flowers $256 \times 256$, where $\alpha$ is uniformly sampled from $[0, \alpha_\mathrm{max}]$ during training.
We sweep $\alpha_\mathrm{max}$ from $80^\circ$ to $90^\circ$ in $2^\circ$ increments.
\cref{tab:abl_max_angle} lists the results at 1, 2 and 4 sampling steps, and $\alpha_\mathrm{max} = 90^\circ$ gives the best generation quality overall.
The metrics improve consistently as $\alpha_\mathrm{max}$ approaches the equator, \eg, the best 1-step FDr$^{6}$ drops from $7.79$ at $80^\circ$ to $2.32$ at $90^\circ$, confirming our hypothesis that the decoder should be trained to reach the equator.
Sampling beyond the training range, $\alpha > \alpha_\mathrm{max}$ (gray rows in \cref{tab:abl_max_angle}), brings no further gain: the performance saturates at its value at $\alpha_\mathrm{max}$.
This is because the decoder is never conditioned on angles $\alpha > \alpha_\mathrm{max}$ during training.

The sampling angle $\alpha$ at inference does not need to equal $\alpha_\mathrm{max}$, because the two play different roles.
$\alpha_\mathrm{max}$ decides which latents the decoder is trained on, while the angle conditioning tells the decoder how to treat a latent.
During training, the decoder learns a different behavior for each angle.
At the equator ($\alpha = 90^\circ$), the latent carries no information about the image, so the outputs concentrate around the mean image of the class.
In the generation regime with $\alpha_\mathrm{cutoff} \leqslant \alpha < 90^\circ$, the latent keeps some clean signal, so the outputs are diverse.
At inference, one-step generation still decodes a random point on the equator, but we condition the decoder on a lower angle, \eg, $\alpha = 85^\circ$.
The decoder then renders the random point as it renders latents around $85^\circ$: as a distinct image rather than the mean image.
The angle embedding thus plays a role similar to a guidance scale that the model is conditioned on during training \citep{meng2023distillation,chen2025visual,geng2026improved,lu2026one}: a scalar seen in training that steers samples away from the average prediction at inference.
This explains why the sampling angle controls the generation diversity (\cref{appsec:abl_sampling_angle}).

\begin{table*}[!t]
    \vspace{-1.5em}
    \centering
    \renewenvironment{table}[1][]{\captionsetup{type=table,position=top}}{}
    \begin{table}[h]
    \centering
    \newcommand{\oob}{\color{gray6}} 
    \caption{
        \textbf{Maximum training angle $\alpha_\mathrm{max}$} of Sphere2-B trained on Oxford Flowers $256 \times 256$.
        $\alpha_\mathrm{max}$ denotes the maximum training angle in \cref{eq:rotation} and $\alpha_\mathrm{max} = 90^\circ$ reaches the equator (default).
        Each block has a different $\alpha_\mathrm{max}$, swept over the sampling angle $\alpha$ at inference, with the best value of each column in the block in bold.
        \textcolor{gray6}{Gray} rows sample beyond the training range, $\alpha > \alpha_\mathrm{max}$, where the performance saturates.
        Columns 1, 2 and 4 give the number of sampling steps $K$.
    }
    \label{tab:abl_max_angle}
    \footnotesize
    \setlength{\tabcolsep}{4.9pt}
    \begin{tabular}{c ccc ccc p{2pt} c ccc ccc}
        \toprule
        \multirow{2}{*}[-0.5ex]{$\alpha$}                                     & \multicolumn{3}{c}{FDr$^{6}$ $\downarrow$ @ $K$ steps} & \multicolumn{3}{c}{gFID $\downarrow$ @ $K$ steps}                     &                    &
        \multirow{2}{*}[-0.5ex]{$\alpha$}                                     & \multicolumn{3}{c}{FDr$^{6}$ $\downarrow$ @ $K$ steps} & \multicolumn{3}{c}{gFID $\downarrow$ @ $K$ steps}                                                                                                                                                                                                                                                                \\
        \cmidrule(lr){2-4} \cmidrule(lr){5-7} \cmidrule(lr){10-12} \cmidrule(lr){13-15}
                                                                              & 1                                                      & 2                                                                     & 4                  & 1                   & 2                   & 4                   &  &                 & 1                  & 2                  & 4                  & 1                   & 2                 & 4                   \\
        \cmidrule{1-7} \cmidrule{9-15}
        \multicolumn{7}{l}{\cellcolor{gray1}$\alpha_\mathrm{max} = 80^\circ$} &                                                        & \multicolumn{7}{l}{\cellcolor{gray1}$\alpha_\mathrm{max} = 82^\circ$}                                                                                                                                                                                                                                            \\
        $70^\circ$                                                            & 10.06                                                  & 7.64                                                                  & 6.84               & 26.41               & 29.56               & 30.86               &  & $70^\circ$      & 9.04               & 7.18               & 6.52               & 27.66               & 30.24             & 30.83               \\
        $72^\circ$                                                            & \ \ 9.09                                               & 6.91                                                                  & 6.20               & 25.75               & 28.66               & 29.22               &  & $72^\circ$      & 8.05               & 6.34               & 5.72               & 24.22               & 27.27             & 27.45               \\
        $74^\circ$                                                            & \ \ 8.80                                               & 6.67                                                                  & 6.04               & 25.50               & 26.96               & 27.72               &  & $74^\circ$      & 7.67               & 6.01               & 5.43               & 23.56               & 25.47             & 25.14               \\
        $76^\circ$                                                            & \ \ 8.48                                               & 6.42                                                                  & 5.79               & 24.30               & 24.57               & 25.12               &  & $76^\circ$      & 7.37               & 5.80               & 5.25               & 22.76               & 23.88             & 23.46               \\
        $78^\circ$                                                            & \ \ 8.08                                               & 6.12                                                                  & 5.53               & 22.19               & 21.85               & 22.43               &  & $78^\circ$      & 7.08               & 5.60               & 5.05               & 21.59               & 21.96             & 21.72               \\
        $80^\circ$                                                            & \ \ 7.80                                               & \textbf{5.93}                                                         & 5.36               & 21.46               & 20.16               & 20.69               &  & $80^\circ$      & 6.67               & 5.24               & 4.73               & 19.74               & 19.54             & 19.24               \\
        \oob $82^\circ$                                                       & \oob \ \ 7.82                                          & \oob 5.95                                                             & \oob 5.36          & \oob 21.47          & \oob 20.37          & \oob 20.76          &  & $82^\circ$      & 6.44               & 5.09               & 4.58               & 18.91               & \textbf{17.93}    & 17.50               \\
        \oob $84^\circ$                                                       & \oob \ \ \textbf{7.79}                                 & \oob \textbf{5.93}                                                    & \oob 5.36          & \oob 21.51          & \oob 20.29          & \oob \textbf{20.46} &  & \oob $84^\circ$ & \oob \textbf{6.43} & \oob 5.09          & \oob \textbf{4.57} & \oob 18.88          & \oob 18.05        & \oob 17.42          \\
        \oob $85^\circ$                                                       & \oob \ \ \textbf{7.79}                                 & \oob 5.95                                                             & \oob 5.37          & \oob \textbf{21.39} & \oob \textbf{20.15} & \oob 20.61          &  & \oob $85^\circ$ & \oob \textbf{6.43} & \oob 5.09          & \oob \textbf{4.57} & \oob 18.78          & \oob 18.03        & \oob \textbf{17.36} \\
        \oob $86^\circ$                                                       & \oob \ \ 7.82                                          & \oob 5.94                                                             & \oob \textbf{5.35} & \oob 21.48          & \oob 20.28          & \oob 20.65          &  & \oob $86^\circ$ & \oob 6.44          & \oob \textbf{5.08} & \oob 4.59          & \oob \textbf{18.74} & \oob 18.07        & \oob 17.51          \\
        \cmidrule{1-7} \cmidrule{9-15}
        \multicolumn{7}{l}{\cellcolor{gray1}$\alpha_\mathrm{max} = 84^\circ$} &                                                        & \multicolumn{7}{l}{\cellcolor{gray1}$\alpha_\mathrm{max} = 86^\circ$}                                                                                                                                                                                                                                            \\
        $70^\circ$                                                            & \ \ 8.03                                               & 6.58                                                                  & 6.11               & 25.12               & 29.63               & 29.83               &  & $70^\circ$      & 5.71               & 4.39               & 4.17               & 15.79               & 18.38             & 19.90               \\
        $72^\circ$                                                            & \ \ 6.89                                               & 5.58                                                                  & 5.17               & 22.97               & 26.20               & 25.97               &  & $72^\circ$      & 4.78               & 3.63               & 3.38               & 14.34               & 15.86             & 15.79               \\
        $74^\circ$                                                            & \ \ 6.59                                               & 5.28                                                                  & 4.85               & 21.81               & 23.90               & 23.26               &  & $74^\circ$      & 4.67               & 3.49               & 3.24               & 13.91               & 14.69             & 14.46               \\
        $76^\circ$                                                            & \ \ 6.33                                               & 5.06                                                                  & 4.62               & 20.34               & 21.64               & 20.88               &  & $76^\circ$      & 4.50               & 3.37               & 3.14               & 13.30               & 13.70             & 13.54               \\
        $78^\circ$                                                            & \ \ 6.02                                               & 4.79                                                                  & 4.37               & 18.55               & 18.92               & 18.22               &  & $78^\circ$      & 4.21               & 3.17               & 2.98               & 12.41               & 12.44             & 12.38               \\
        $80^\circ$                                                            & \ \ 5.65                                               & 4.45                                                                  & 4.05               & 17.11               & 16.84               & 16.12               &  & $80^\circ$      & 3.94               & 3.02               & 2.82               & 11.47               & 11.31             & 11.16               \\
        $82^\circ$                                                            & \ \ 5.35                                               & 4.24                                                                  & 3.83               & 15.80               & 15.22               & 14.46               &  & $82^\circ$      & 3.69               & 2.87               & 2.71               & 11.19               & 10.83             & 10.70               \\
        $84^\circ$                                                            & \ \ \textbf{5.02}                                      & \textbf{3.98}                                                         & 3.63               & \textbf{14.53}      & \textbf{13.47}      & 12.98               &  & $84^\circ$      & 3.44               & 2.72               & \textbf{2.56}      & \textbf{10.87}      & 10.28             & \ \ \textbf{9.98}   \\
        \oob $85^\circ$                                                       & \oob \ \ \textbf{5.02}                                 & \oob 3.99                                                             & \oob \textbf{3.62} & \oob 14.64          & \oob 13.67          & \oob 12.85          &  & $85^\circ$      & 3.36               & \textbf{2.70}      & \textbf{2.56}      & 11.08               & \textbf{10.24}    & 10.00               \\
        \oob $86^\circ$                                                       & \oob \ \ \textbf{5.02}                                 & \oob 4.00                                                             & \oob \textbf{3.62} & \oob 14.64          & \oob 13.56          & \oob \textbf{12.83} &  & $86^\circ$      & \textbf{3.34}      & 2.72               & \textbf{2.56}      & 11.62               & 10.55             & 10.12               \\
        \cmidrule{1-7} \cmidrule{9-15}
        \multicolumn{7}{l}{\cellcolor{gray1}$\alpha_\mathrm{max} = 88^\circ$} &                                                        & \multicolumn{7}{l}{\cellcolor{gray1}$\alpha_\mathrm{max} = 90^\circ$}                                                                                                                                                                                                                                            \\
        $70^\circ$                                                            & \ \ 4.86                                               & 3.64                                                                  & 3.52               & 14.84               & 17.32               & 18.78               &  & $70^\circ$      & 3.95               & 3.23               & 3.29               & 14.00               & 15.92             & 17.57               \\
        $72^\circ$                                                            & \ \ 3.79                                               & 2.82                                                                  & 2.76               & 12.94               & 13.92               & 14.11               &  & $72^\circ$      & 2.91               & 2.43               & 2.49               & 11.90               & 12.71             & 13.22               \\
        $74^\circ$                                                            & \ \ 3.40                                               & 2.56                                                                  & 2.52               & 12.23               & 12.53               & 12.20               &  & $74^\circ$      & 2.67               & 2.22               & 2.26               & 10.85               & 11.11             & 11.18               \\
        $76^\circ$                                                            & \ \ 3.26                                               & 2.47                                                                  & 2.45               & 11.40               & 11.43               & 11.28               &  & $76^\circ$      & 2.60               & 2.16               & 2.23               & \ \ 9.99            & \ \ 9.97          & 10.24               \\
        $78^\circ$                                                            & \ \ 3.00                                               & 2.32                                                                  & 2.30               & 10.46               & 10.36               & 10.43               &  & $78^\circ$      & 2.53               & 2.13               & 2.19               & \ \ 9.69            & \ \ 9.57          & \ \ 9.67            \\
        $80^\circ$                                                            & \ \ 2.87                                               & 2.25                                                                  & 2.23               & 10.05               & \ \ 9.76            & \ \ 9.66            &  & $80^\circ$      & 2.46               & 2.10               & 2.17               & \ \ 9.05            & \ \ 8.89          & \ \ 9.11            \\
        $82^\circ$                                                            & \ \ 2.69                                               & 2.14                                                                  & 2.12               & \ \ 9.73            & \ \ 9.18            & \ \ 9.12            &  & $82^\circ$      & 2.38               & 2.06               & 2.10               & \ \ \textbf{8.85}   & \ \ 8.60          & \ \ 8.78            \\
        $84^\circ$                                                            & \ \ 2.52                                               & 2.07                                                                  & 2.04               & \ \ \textbf{9.64}   & \ \ \textbf{8.84}   & \ \ \textbf{8.73}   &  & $84^\circ$      & 2.33               & \textbf{2.03}      & \textbf{2.04}      & \ \ 9.06            & \ \ \textbf{8.49} & \ \ \textbf{8.50}   \\
        $85^\circ$                                                            & \ \ \textbf{2.47}                                      & \textbf{2.06}                                                         & \textbf{2.01}      & 10.14               & \ \ 9.27            & \ \ 9.04            &  & $85^\circ$      & \textbf{2.32}      & 2.06               & 2.06               & \ \ 9.70            & \ \ 9.07          & \ \ 8.83            \\
        $86^\circ$                                                            & \ \ 2.50                                               & 2.12                                                                  & 2.06               & 11.07               & \ \ 9.99            & \ \ 9.62            &  & $86^\circ$      & 2.41               & 2.15               & 2.12               & 10.40               & \ \ 9.65          & \ \ 9.38            \\
        \bottomrule
    \end{tabular}
\end{table}

    \vspace{2.45em}
    \begin{table}[h]
    \centering
    \caption{
        \textbf{Component roadmap} of Sphere2-B trained on Oxford Flowers $256 \times 256$.
        The first row is a baseline with a frozen encoder $\mathcal{E}$, trained with the pixel and LPIPS losses at all noise angles.
        Each later row adds one component on top of the row above it.
        $\alpha_\mathrm{cutoff}$ restricts the pixel and LPIPS losses to angles below the cutoff, and applies $\cL_\mathrm{sem}$ to angles above it.
        The \shaded{shaded} last row is the default Sphere2-B.
        Columns 1, 2 and 4 give the number of sampling steps.
        $\cL_\mathrm{FD\text{-}lite}$ is not part of this ablation: it is used only for the Sphere2$^\dagger$ models on ImageNet (\cref{appsec:distribution_shift}).
        The numbers of row (e) differ slightly from \cref{tab:fnl_flowers}, since it is trained separately from the main model to decompose the losses one by one, and training has some stochasticity.
    }
    \label{tab:abl_component_roadmap}
    \footnotesize
    \setlength{\tabcolsep}{5pt}
    \begin{tabular}{c l ccc ccc}
        \toprule
                                                                 &
        \multirow{2}{*}[-0.5ex]{Component}                       &
        \multicolumn{3}{c}{{FDr$^{6}$} $\downarrow$ @ $K$ steps} &
        \multicolumn{3}{c}{{gFID} $\downarrow$ @ $K$ steps}                                                                                                                                                        \\
        \cmidrule(lr){3-5} \cmidrule(lr){6-8}
                                                                 &                                                    & 1       & 2       & 4       & 1                  & 2                  & 4                  \\
        \midrule
        (a)                                                      & frozen encoder $\mathcal{E}$                       & 51.93   & 31.84   & 21.45   & 332.27             & 202.51             & 119.53             \\
        (b)                                                      & + angle conditioning                               & 35.65   & 24.69   & 20.11   & 219.34             & 136.37             & 103.29             \\
        (c)                                                      & + $\cL_\mathrm{sem}$ with $\alpha_\mathrm{cutoff}$ & \; 5.39 & \; 4.47 & \; 4.38 & \; 15.26           & \; 14.41           & \; 14.71           \\
        (d)                                                      & + $\cL_\mathrm{score}$                             & \; 2.63 & \; 2.26 & \; 2.22 & \; 10.01           & \; \phantom{0}9.64 & \; \phantom{0}9.56 \\
        \rowcolor{gray1}
        (e)                                                      & + $\cL_\mathrm{lat\text{-}con}$                    & \; 2.33 & \; 2.01 & \; 1.98 & \; \phantom{0}9.32 & \; \phantom{0}8.79 & \; \phantom{0}8.74 \\
        \bottomrule
    \end{tabular}
\end{table}

\end{table*}

\subsection{Component Roadmap}
\label{appsec:abl_component_roadmap}

\begin{figure}[h]
    \centering
    \includegraphics[width=1.0\linewidth]{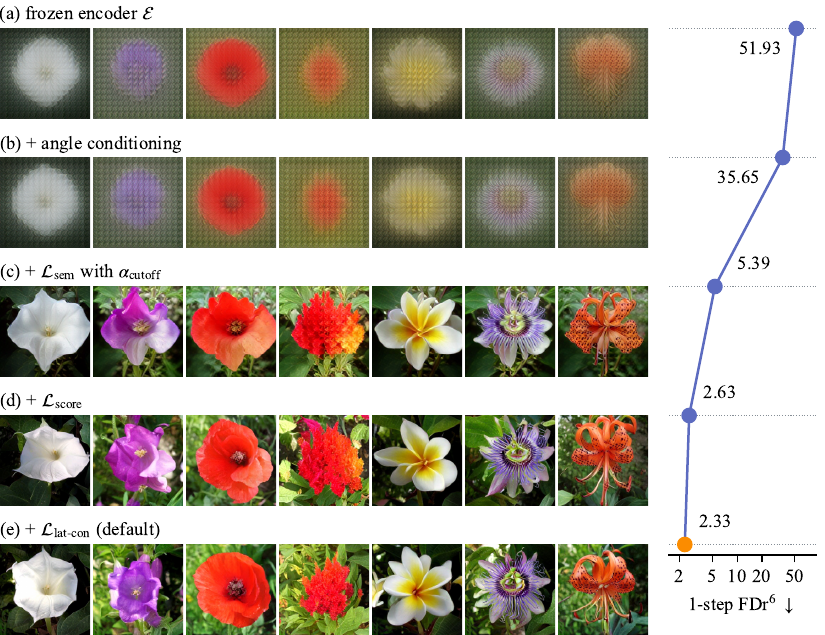}
    \caption{
        \textbf{Component roadmap in samples.}
        Left: 1-step samples of Sphere2-B on Oxford Flowers $256 \times 256$ for each row of \cref{tab:abl_component_roadmap}, with the same classes and sampling seed in every block.
        Right: the 1-step FDr$^{6}$ of each row on a log axis; the orange point is the default model.
    }
    \label{fig:abl_component_roadmap_vis}
\end{figure}

\cref{tab:abl_component_roadmap} adds the components of Sphere Encoder 2 one at a time with Sphere2-B on Oxford Flowers $256 \times 256$, in the order (a) frozen encoder $\cE$, (b) angle conditioning, (c) angle cutoff with $\cL_\mathrm{sem}$, (d) $\cL_\mathrm{score}$, and (e) $\cL_\mathrm{lat\text{-}con}$.
The baseline (a) applies the pixel-wise reconstruction loss along the whole arc and generates poorly, since this loss is minimized by the mean of all plausible images (\cref{sec:sphere2}).
Angle conditioning (b) helps but does not solve the problem; its effect on the full model is ablated separately in \cref{tab:abl_angle_cond}.
The angle cutoff (c) brings by far the largest gain, cutting 1-step FDr$^{6}$ from $35.65$ to $5.39$: it confines the pixel-wise loss to the reconstruction regime and trains the generation regime with $\cL_\mathrm{sem}$ instead.
$\cL_\mathrm{score}$ (d) then halves FDr$^{6}$ to $2.63$, confirming its effectiveness to improve the generation fidelity, and $\cL_\mathrm{lat\text{-}con}$ (e) refines it to $2.33$.
Every component improves both metrics at every step count, and the last row is the default recipe of Sphere2.
\cref{fig:abl_component_roadmap_vis} shows 1-step samples of each row: (a) and (b) generate blurry mean images, and sharp flowers appear once the angle cutoff is added in (c).

\subsection{Angle Conditioning}

\noindent
\begin{minipage}[c]{0.475\linewidth}
    $\cL_\mathrm{score}$ asks the decoder to cover the real distribution rather than collapse onto its mean.
    We hypothesize that an angle-blind decoder hedges toward the mean, since it cannot tell latents at different angles apart.
    \cref{tab:abl_angle_cond} ablates conditioning the decoder on $\alpha$ -- finding our hypothesis is correct.
    Angle conditioning effectively improves metrics, so we keep it by default.
\end{minipage}
\hfill
\begin{minipage}[c]{0.475\linewidth}
    \begin{center}
    \small
    \setlength{\tabcolsep}{3.9pt}
    \captionof{table}{
        \textbf{Angle conditioning} in the Sphere2-B trained on Oxford Flowers $256 \times 256$.
    }
    \vspace{-0.5em}
    \begin{tabular}{c ccc ccc}
        \toprule
        \multirow{2}{*}[-0.5ex]{$\alpha$ cond}       &
        \multicolumn{3}{c}{{FDr$^{6}$} $\downarrow$} &
        \multicolumn{3}{c}{{gFID} $\downarrow$}                                                                                                               \\
        \cmidrule(lr){2-4} \cmidrule(lr){5-7}
                                                     & 1             & 2             & 4             & 1              & 2                 & 4                 \\
        \midrule
        -                                            & 4.00          & 3.21          & 3.23          & 14.02          & 12.09             & 11.95             \\
        \checkmark                                   & \textbf{3.92} & \textbf{3.14} & \textbf{3.04} & \textbf{10.29} & \ \ \textbf{8.24} & \ \ \textbf{7.51} \\
        \bottomrule
    \end{tabular}
    \label{tab:abl_angle_cond}
\end{center}

\end{minipage}
\vspace{0.5em}

This ablation, as well as those in \cref{tab:abl_latent,tab:abl_score_sigma_distribution}, was run early, before the cutoff sweep, with $\alpha_\mathrm{cutoff} = 80^\circ$ instead of the default $70^\circ$, and the same bandwidth $\beta = 5^\circ$.
All rows within each table share this setting, so the comparisons hold.

\clearpage
\vspace*{\dimexpr-\topskip-\parskip-1.8ex-1.5em\relax}
\subsection{Sampling Angle at Inference}
\label{appsec:abl_sampling_angle}

\begin{figure}[h]
    \centering
    \includegraphics[width=1.0\linewidth]{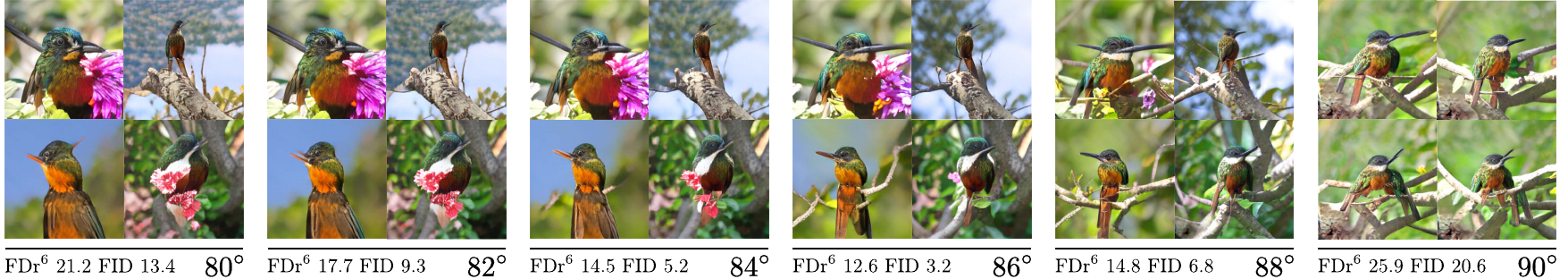}
    \caption{
        \textbf{Sampling angle $\alpha$} with 1-step generation using Sphere2-B trained on ImageNet $256 \times 256$.
        Sampling angle decides the generation diversity by forcing the decoder away from the mean.
    }
    \label{fig:abl_sampling_angle_vis}
\end{figure}

\noindent
\begin{minipage}[c]{0.52\linewidth}
    Since the decoder is conditioned on the angle $\alpha$, we investigate sampling properties at different angles.
    We sweep $\alpha$ from $0^\circ$ to $90^\circ$ in 5-degree increments for one-step generation with Sphere2-B$^\dagger$ trained on ImageNet $256 \times 256$.
    \cref{fig:abl_sampling_angle_vis} shows that the sampling angle controls the diversity of generated images.
    With $\alpha=90^\circ$, the generated images are almost the same, which is expected: the decoder sees only the pure noise $\cF(\be)$ in \cref{eq:rotation} without any reference information from the input image during training.
    Thus, the decoder collapses to the mean of all plausible images in the class.
    With $\alpha < 90^\circ$, the decoder sees a mixture of the clean latent and the noise during training, so it starts to generate images with more diversity.
    \cref{fig:abl_sampling_angle_res} quantifies the effect of sampling angle on ImageNet $256 \times 256$.
\end{minipage}
\hfill
\begin{minipage}[c]{0.43\linewidth}
    \centering
    \includegraphics[width=1.0\linewidth]{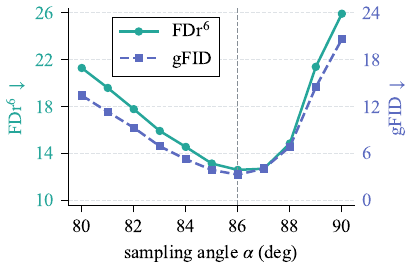}
    \vspace{-1.5em}
    \captionof{figure}{
        \textbf{Sampling angle $\alpha$} with 1-step generation using Sphere2-B$^\dagger$ trained on ImageNet $256 \times 256$.
        FDr$^{6}$ and FID agree in a ``U'' shape along the sampling angle.
    }
    \label{fig:abl_sampling_angle_res}
\end{minipage}
\vspace{0.5em}

For multi-step generation, we apply a dense sweep and investigate the effect of sampling angle $\alpha$ on ImageNet and Oxford Flowers at both resolutions of $256 \times 256$ and $512 \times 512$.
The alpha $\alpha$ is fixed for all intermediate steps in the loop of \cref{alg:forward}.
We found the FDr$^{6}$ and FID do not agree on the best $\alpha$ for multi-step generation.
For simplicity, we pick $\alpha = 85.5^\circ$ for Sphere2-B and $\alpha=84^\circ$ for Sphere2-L to report the main results in \cref{tab:fnl_imagenet,tab:fnl_flowers}, in order to prioritize FDr$^{6}$ over FID.

\subsection{Sweeping Sampling Steps}
\label{appsec:abl_sweep_sampling_steps}

\cref{tab:abl_sweep_sampling_steps} sweeps the number of sampling steps from 1 to 500 for three pixel-space base models on ImageNet $256 \times 256$.
Each model is pushed away from the budget it was designed for: the many-step diffusion model JiT-B (50 steps) toward few steps, and the one-step model pMF-B and the few-step Sphere2-B (1 to 4 steps) toward many steps.

\begin{figure}[h]
    \centering
    \includegraphics[width=1.0\linewidth]{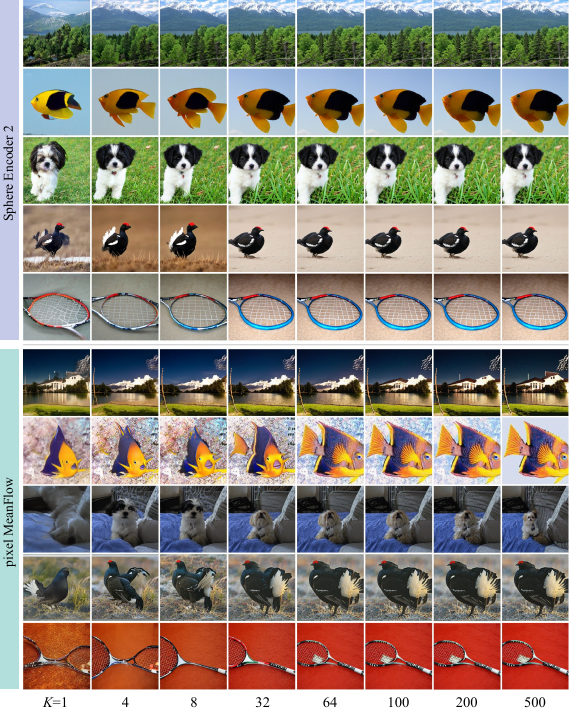}
    \caption{
        \textbf{Sweeping sampling steps $K$} on ImageNet $512 \times 512$ with pMF-L and Sphere2-L.
    }
    \label{fig:abl_sweep_sampling_steps}
\end{figure}

Three observations follow.
First, each model is best near its design budget: JiT-B collapses below 8 steps and plateaus from 32 steps on, with FDr$^{6}$ staying between $15.5$ and $15.9$, pMF-B peaks at 8 steps, and Sphere2-B peaks at 4.
Second, Sphere2-B degrades gracefully when pushed to many steps: its FDr$^{6}$ rises from $8.62$ at 4 steps to $10.05$ at 500 and then saturates, so the refinement loop of \cref{alg:forward} stays stable rather than drifting away.
Third, from 4 steps onward Sphere2-B has the best FDr$^{6}$ of the three models at every step count without any guidance, and its 4-step result is better than the best result of both JiT-B and pMF-B.
\cref{fig:abl_sweep_sampling_steps} shows the same trend qualitatively at $512 \times 512$.

\begin{table}[h]
    \centering
    \caption{
        \textbf{Sweeping sampling steps} $K$ on ImageNet $256 \times 256$ with pixel-space base models.
        CFG doubles NFE, and CFG$^\ast$ is folded into the training target, as in \cref{tab:fnl_imagenet}.
        The qualitative results are in \cref{fig:abl_sweep_sampling_steps}, showing that Sphere2 maintains high fidelity with more than 100 sampling steps.
    }
    \label{tab:abl_sweep_sampling_steps}
    \small
    \setlength{\tabcolsep}{5.7pt}
    \begin{tabular}{l c c cccccccc}
        \toprule
        \multirow{2}{*}[-0.5ex]{Model}  &
        \multirow{2}{*}[-0.5ex]{Params} &
        \multirow{2}{*}[-0.5ex]{Guid.}  &
        \multicolumn{8}{c}{FDr$^{6}$ $\downarrow$ @ $K$ steps}                                                                             \\
        \cmidrule(lr){4-11}
                                        &          &               & 1        & 4        & 8        & 32       & 64    & 100   & 200 & 500 \\
        \midrule
        JiT-B \citeyearpar{li2026back}  & 131M     & CFG
                                        & 222.89   & 45.84         & 23.16    & 15.89    & 15.51    & 15.79    & 15.83 & 15.82             \\
        pMF-B \citeyearpar{lu2026one}   & 118M     & \; CFG$^\ast$
                                        & \; 13.83 & 10.69         & 10.61    & 12.17    & 12.69    & 12.82    & 12.96 & 13.06             \\
        \textbf{Sphere2-B}              & 161M     & --
                                        & \; 14.50 & \ \ 8.62      & \ \ 8.96 & \ \ 9.71 & \ \ 9.91 & \ \ 9.96 & 10.00 & 10.05             \\
        \bottomrule
    \end{tabular}
\end{table}

\subsection{Frozen Latent Space}

\cref{tab:abl_latent} ablates the choice of the frozen encoder $\cE$.
Since the encoder runs at every sampling step, we restrict the comparison to the small and base sizes of two model families, DINOv3 \citep{simeoni2025dinov3} and SigLIP2 \citep{tschannen2025siglip} for training and inference efficiency.
The small size has latent dimension $D = 384$ and the base size $D = 768$.
We also test a lower-dimensional variant of each, adding a single linear layer that reduces $D$ to $256$ for small and $384$ for base.
The results show: (1) DINOv3 outperforms SigLIP2 at every size with original dimensions, and (2) DINOv3-small+ with its original $D = 384$ gives the best trade-off between quality and efficiency.
We therefore use it by default.

\begin{table}[h]
    \centering
    \caption{
        \textbf{Frozen latent space} of the encoder $\cE$ with different pretrained encoders and latent dimensions on Oxford-Flowers $256 \times 256$.
    }
    \label{tab:abl_latent}
    \small
    \setlength{\tabcolsep}{7.7pt}
    \begin{tabular}{l c c ccc ccc}
        \toprule
        \multirow{2}{*}[-0.5ex]{Encoder}             &
        \multirow{2}{*}[-0.5ex]{$d = N \times D$}    &
        \multirow{2}{*}[-0.5ex]{\shortstack{Compression                                                                                                                            \\ ratio}}     &
        \multicolumn{3}{c}{{FDr$^{6}$} $\downarrow$} &
        \multicolumn{3}{c}{{gFID} $\downarrow$}                                                                                                                                    \\
        \cmidrule(lr){4-6} \cmidrule(lr){7-9}
                                                     &                 &   & 1             & 2             & 4             & 1              & 2                & 4                 \\
        \midrule
        DINOv3-small+                                & $16^2\times384$ & 2 & \textbf{3.92} & \textbf{3.14} & \textbf{3.04} & \textbf{10.29} & \; \textbf{8.24} & \ \ \textbf{7.51} \\
        DINOv3-small+                                & $16^2\times256$ & 3 & 4.53          & 3.77          & 3.61          & 17.67          & 13.64            & 12.27             \\
        \cmidrule(lr){1-9}
        DINOv3-base                                  & $16^2\times768$ & 1 & 4.28          & 3.58          & 3.51          & 11.47          & \; 9.28          & \ \ 8.69          \\
        DINOv3-base                                  & $16^2\times384$ & 2 & 5.17          & 4.48          & 4.46          & 18.18          & 15.67            & 15.78             \\
        SigLIP2-base                                 & $16^2\times768$ & 1 & 5.82          & 4.37          & 4.18          & 15.15          & 13.73            & 15.15             \\
        SigLIP2-base                                 & $16^2\times384$ & 2 & 4.91          & 4.25          & 4.24          & 16.32          & 13.15            & 13.07             \\
        \bottomrule
    \end{tabular}
\end{table}

\subsection{Score Feature Space}

$\cL_\mathrm{score}$ needs a frozen feature space $\phi$ and a choice of layers.
We compare two lightweight extractors: ConvNeXt V2-N, our default, and DINO-S/8, the classic choice for GAN losses \citep{zheng2026diffusion}.
For each, we sweep which layers feed the score models in \cref{tab:abl_score_feature_layers}, using Sphere2-B.
Two layers are enough for both extractors: a middle and a deep block for DINO-S/8, and the two middle stages for ConvNeXt V2-N.
Adding more layers costs up to twice the parameters and helps a little.

\begin{table}[h]
    \centering
    \caption{
        \textbf{Feature extractor $\phi$ and its layers used by the score loss} on Oxford Flowers $256 \times 256$.
        Each score model is a two-layer transformer with width 384.
        Params counts total parameters of all score models.
        Layers are 0-indexed: transformer blocks for DINO-S/8 and stages for ConvNeXt V2-N.
        Params of DINO-S/8 is 21.67M, and ConvNeXt V2-N \texttt{fcmae} is 14.98M.
    }
    \label{tab:abl_score_feature_layers}
    \small
    \setlength{\tabcolsep}{5pt}
    \begin{tabular}{c l c cc cc}
        \toprule
        \multirow{2}{*}[-0.5ex]{Feature $\phi$}      &
        \multirow{2}{*}[-0.5ex]{Layers}              &
        \multirow{2}{*}[-0.5ex]{Params}              &
        \multicolumn{2}{c}{{FDr$^{6}$} $\downarrow$} &
        \multicolumn{2}{c}{{gFID} $\downarrow$}                                                                                            \\
        \cmidrule(lr){4-5} \cmidrule(lr){6-7}
                                                     &          &        & 1             & 4             & 1              & 4              \\
        \midrule
        \cellcolor{white}
        \multirow{4}{*}{DINO-S/8}
                                                     & 5 11     & 25.19M & \textbf{6.16} & \textbf{4.63} & \textbf{24.86} & \textbf{17.17} \\
                                                     & 3 7 11   & 37.79M & 5.95          & 4.49          & 27.23          & 17.68          \\
                                                     & 5 8 11   & 37.79M & 6.20          & 4.72          & 24.28          & 17.80          \\
                                                     & 2 5 8 11 & 50.38M & 6.48          & 5.12          & 27.20          & 20.28          \\
        \cmidrule(lr){1-7}
        \cellcolor{white}
        \multirow{2}{*}{\shortstack{ConvNeXt                                                                                               \\V2-N                                                                                         }}
                                                     & 1 2      & 24.16M & \textbf{6.51} & \textbf{4.88} & \textbf{26.80} & \textbf{18.60} \\
                                                     & 1 3      & 44.79M & 6.42          & 5.18          & 29.85          & 21.52          \\
        \bottomrule
    \end{tabular}
\end{table}

\subsection{Score Sigma Distribution}

\noindent
\begin{minipage}[c]{0.50\linewidth}
    $\cL_\mathrm{score}$ interleaves two procedures: training the score models and transporting the generator with them.
    The score models are standard EDM denoisers, trained on the noisy features of \cref{eq:score_matching_noisy_feature} with $\sigma$ drawn from the EDM log-normal $p(\sigma)$, $(p_\mathrm{mean}, p_\mathrm{std}) = (-1.2, 1.2)$, without clamping.
    For the transport in \cref{eq:score_matching_transport}, we draw $\sigma$ from a second log-normal $q(\sigma)$ with $(q_\mathrm{mean}, q_\mathrm{std})$.
    Two considerations guide the choice of $q(\sigma)$.
    First, $q(\sigma)$ should stay within the bulk of $p(\sigma)$, so the score models are never queried at noise levels they were not trained on.
    Second, $q(\sigma)$ should lean toward small noise levels, where the score difference carries the fine detail that pushes the generator toward high-fidelity and sharp generations.

\end{minipage}
\hfill
\begin{minipage}[c]{0.45\linewidth}
    \centering
    \includegraphics[width=0.95\linewidth]{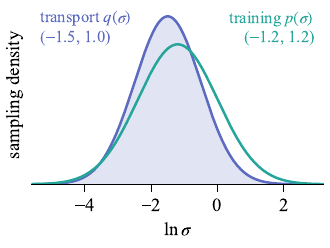}
    \vspace{-0.5em}
    \captionof{figure}{
        \textbf{Noise sampling distribution for score models} of $\cL_\mathrm{score}$ in training $p(\sigma)$ and transport $q(\sigma)$.
    }
    \label{fig:abl_score_sigma_distribution}
\end{minipage}
\vspace{0.5em}

\cref{tab:abl_score_sigma_distribution} sweeps $(q_\mathrm{mean}, q_\mathrm{std})$ with $p(\sigma)$ fixed, using Sphere2-B.
The best $q(\sigma)$ is $(-1.5, 1.0)$: shifted toward small $\sigma$ and slightly narrower than $p(\sigma)$, as \cref{fig:abl_score_sigma_distribution} shows.
Pushing further hurts.
At $(-1.8, 0.8)$, $q(\sigma)$ concentrates away from the mode of $p(\sigma)$, and FDr$^{6}$ becomes worse.
Additionally, the ImageNet finetuned version of ConvNeXt V2-N \texttt{ft-in1k} provides a better feature space than the pretrained ConvNeXt V2-N \texttt{fcmae} on Oxford Flowers, so we use it by default for all experiments.

\begin{table}[h]
    \centering
    \caption{
        \textbf{Sigma distribution for transport procedure} of the score loss on Flowers $256 \times 256$.
    }
    \label{tab:abl_score_sigma_distribution}
    \small
    \setlength{\tabcolsep}{8pt}
    \begin{tabular}{c c cc cc}
        \toprule
        \multirow{2}{*}[-0.5ex]{Feat. $\phi$}                                                                                                                                                                        &
        \multirow{2}{*}[-0.5ex]{\shortstack{Transport                                                                                    \\{\scriptsize $(q_\mathrm{mean}, q_\mathrm{std})$}}} &
        \multicolumn{2}{c}{{FDr$^{6}$} $\downarrow$}                                                                                                                                                                 &
        \multicolumn{2}{c}{{gFID} $\downarrow$}                                                                                                                                                                                                                                                          \\
        \cmidrule(lr){3-4} \cmidrule(lr){5-6}
                                                                                                                                                                                                                     &               & 1             & 4             & 1              & 4                \\
        \midrule
        \multirow{4}{*}{DINO-S/8}                                                                                                                                                                                    & $(-1.2, 1.0)$ & 4.54          & 3.93          & 13.69          & 10.89            \\
                                                                                                                                                                                                                     & $(-1.5, 1.0)$ & 4.57          & 3.94          & 13.35          & 10.54            \\
                                                                                                                                                                                                                     & $(-1.8, 1.0)$ & 4.61          & 3.96          & 13.19          & 10.37            \\
                                                                                                                                                                                                                     & $(-1.8, 0.8)$ & 4.97          & 4.28          & 14.46          & 11.02            \\
        \cmidrule(lr){1-6}
                                                                                                                                                                                                                     & $(-1.2, 1.0)$ & 3.82          & 3.09          & 11.40          & \; 8.28          \\
        \cellcolor{white}                                                                                                                                                                                            & $(-1.5, 1.0)$ & \textbf{3.92} & \textbf{3.04} & \textbf{10.29} & \; \textbf{7.51} \\
                                                                                                                                                                                                                     & $(-1.8, 1.0)$ & 4.81          & 3.64          & 13.50          & \; 9.40          \\
        \multirow{-4}{*}{\shortstack{ConvNeXt                                                                                                                                                                                                                                                            \\V2-N\\\texttt{ft-in1k}}}
                                                                                                                                                                                                                     & $(-1.8, 0.8)$ & 6.16          & 5.04          & 15.78          & 13.90            \\
        \cmidrule(lr){1-6}
        \multirow{4}{*}{\shortstack{ConvNeXt                                                                                                                                                                                                                                                             \\V2-N\\\texttt{fcmae}}}
                                                                                                                                                                                                                     & $(-1.2, 1.0)$ & 5.03          & 3.83          & 15.07          & 13.09            \\
                                                                                                                                                                                                                     & $(-1.5, 1.0)$ & 4.92          & 3.96          & 14.00          & 12.58            \\
                                                                                                                                                                                                                     & $(-1.8, 1.0)$ & 5.15          & 4.18          & 14.74          & 12.85            \\
                                                                                                                                                                                                                     & $(-1.8, 0.8)$ & 5.48          & 4.21          & 14.37          & 13.22            \\
        \bottomrule
    \end{tabular}
\end{table}

\section{Text-to-Image Generation on Oxford Flowers}
\label{appsec:t2i_flowers}

In this section, we build a text-to-image version of Sphere Encoder 2 on Oxford Flowers $256 \times 256$ to show that it generalizes beyond class-conditional generation.
The model and all training hyperparameters stay the same as in the class-conditional setting.
Only three components change: the training data, the decoder condition, and the two-layer score models of $\cL_\mathrm{score}$ condition.

\textbf{Captions.}
Oxford Flowers comes with class labels but no captions, so we caption its 8,189 images with \texttt{Qwen/Qwen2.5-VL-7B-Instruct} \citep{qwen2.5-VL}.
Each image has 20 captions, from short (5 words) to long (40 words), so the training text covers different levels of detail.

\textbf{Decoder condition.}
We encode each caption with the SigLIP2 text tower \citep{tschannen2025siglip}, which takes at most 64 tokens, and project the token embeddings to the decoder hidden size with RMSNorm $\to$ Linear $\to$ GELU $\to$ Linear.
Following Qwen-Image-2.1 \citep{qwen2026qwenimage21} and FLUX 3 Image Klein \citep{bfl2026flux3image}, we prepend the projected text tokens $\bt$ to the image tokens $\hat{\z}(\alpha)$ as in-context tokens, so the decoder conditions on the text through self-attention.
The angle $\alpha$ is still injected through adaLN-Zero as in the class-conditional model, now with the angle embedding alone, since there is no class embedding to add it to.
The text tokens thus take the place of the class embedding $\y$, and the conditional generation of \cref{eq:generation_conditional} becomes
\begin{align}
    \hat{\X} = \cD(\cF(\be), \bt, \alpha).
\end{align}

\textbf{Score model condition.}
The score models of $\cL_\mathrm{score}$ also take $\y$ through adaLN-Zero, and we replace it with the pooled output of the SigLIP2 text tower.

\begin{figure}[h]
    \centering
    \includegraphics[width=1.0\linewidth]{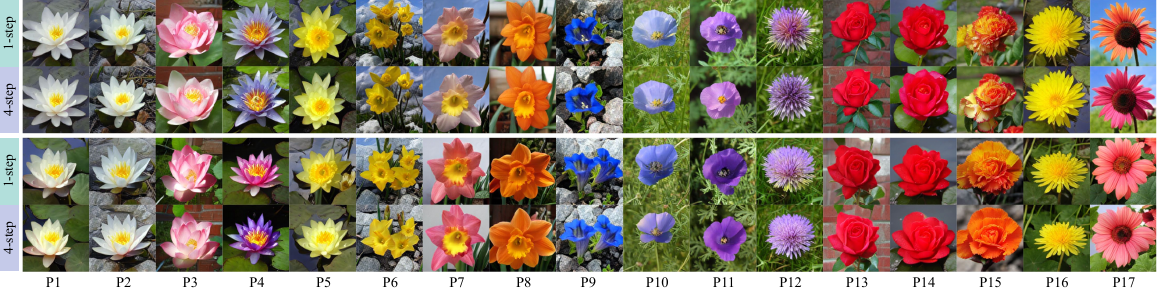}
    \caption{
        \textbf{Uncurated text-to-image generation} on Oxford Flowers $256 \times 256$ by Sphere2-B with 1 and 4 sampling steps.
        Each column is one prompt from \cref{tab:t2i_flowers_prompts}.
        Best viewed zoomed in.
    }
    \label{fig:abl_t2i_flowers}
    \vspace{1.5em}
    \centering
\captionsetup{position=top}
\captionof{table}{
      \textbf{Text prompts} for text-to-image generation on Oxford Flowers $256 \times 256$ in \cref{fig:abl_t2i_flowers}.
      All prompts are unseen and out of distribution: none appears among the training captions, and each describes a flower outside the training data.
}
\label{tab:t2i_flowers_prompts}
\small
\setlength{\tabcolsep}{4.2pt}
\renewcommand{\arraystretch}{1.1}
\newcolumntype{Y}[1]{>{\hsize=#1\hsize\raggedright\arraybackslash\hangindent=2.2em\hangafter=1}X}
\begin{tabularx}{\linewidth}{Y{1.15} @{\hspace{1.5em}} Y{0.85}}
      \toprule
      P1: \textit{A white water lily floating on a pond among lily pads.} & P10: \textit{A blue poppy in a field.}               \\
      P2: \textit{A white water lily growing between stones.}             & P11: \textit{A purple poppy in a field.}             \\
      P3: \textit{A pink lotus flower in front of a red brick wall.}      & P12: \textit{A purple dandelion in the green grass.} \\
      P4: \textit{A blue water lily floating on a pond.}                  & P13: \textit{A red rose in front of a brick wall.}   \\
      P5: \textit{Yellow daffodils floating among lily pads.}             & P14: \textit{A red rose floating on a pond.}         \\
      P6: \textit{Yellow daffodils among rocks on a mountain slope.}      & P15: \textit{A dandelion-like red rose.}             \\
      P7: \textit{A pink daffodil with a trumpet-shaped center.}          & P16: \textit{A yellow dandelion floating on the}     \\
      P8: \textit{An orange daffodil with a trumpet-shaped center.}       & \phantom{P16:} \textit{surface of a pond.}           \\
      P9: \textit{A blue gentian among rocks on a mountain slope.}        & P17: \textit{A pink sunflower with a dark center.}   \\
      \bottomrule
\end{tabularx}

\end{figure}

\cref{tab:t2i_flowers_prompts} lists the prompts we use for text-to-image generation in \cref{fig:abl_t2i_flowers} with Sphere2-B at 1 and 4 sampling steps.
All prompts are out of distribution: each asks for a scene, color, or shape that the training images do not show, such as a water lily growing between stones (P2), a dandelion-like rose (P15), or a pink sunflower, although sunflowers are always yellow in the training set (P17).
With only the three changes above, Sphere2-B follows these unseen prompts already at 1 step, rendering colors and shapes that the training images do not show, such as a blue water lily (P4), a purple dandelion (P12), and a pink sunflower (P17).
Text conditioning thus fits into Sphere Encoder 2 without changing its training recipe, and we leave scaling it to large text-image datasets and higher resolutions to future work.

\section{Score Matching Loss}
\label{appsec:score_matching}

\subsection{Differentiable Augmentation}

We apply differentiable augmentations \citep{zhao2020differentiable} to the decoder output before it is passed to the frozen feature extractor $\phi$ during score model training.
The decoder produces images of size $256 \times 256$ or $512 \times 512$, and we mainly investigate DINO-S/8 \citep{caron2021emerging} and ConvNeXt V2-N(ano) \citep{woo2023convnext} as the feature extractor.
For DINO-S/8, instead of resizing the image to $224 \times 224$, we interpolate the positional embedding to match the image size.
We use two differentiable augmentations: horizontal flip with probability 0.5, and translation with probability 1.0.
The translation ratio is $0.015625$ for both resolutions, i.e., up to 4 pixels at $256 \times 256$ and 8 pixels at $512 \times 512$, with reflect padding.
The translation smooths the patch boundaries.
Since DINO-S/8 uses $8 \times 8$ patches, a shift of up to 4 or 8 pixels is half a patch at $256 \times 256$ and a full patch at $512 \times 512$, so the extractor sees the generated image at every offset relative to its patch grid.

\subsection{Feature Standardization}

\cref{eq:score_matching_clean_feature} standardizes the output of each scored layer $j$ of $\phi$.
Since the procedure is identical for every layer, we drop the index $j$ here.
The raw feature $\phi(\X) \in \real^{L \times C}$ has $L$ spatial locations (ConvNeXt V2-N) or tokens (DINO-S/8) and $C$ channels.
EDM \citep{karras2022elucidating} assumes the input has zero mean and standard deviation $\sigma_\mathrm{data} = 0.5$, so we standardize the layer feature to this scale before it enters the score models.

Given a global batch of $B$ real images, we compute the per-channel mean and second moment of the raw features,
\begin{align}
    m = \frac{1}{BL} \sum_{b=1}^{B} \sum_{l=1}^{L} \phi(\X_b)^{l},
    \quad
    M = \frac{1}{BL} \sum_{b=1}^{B} \sum_{l=1}^{L} \big( \phi(\X_b)^{l} \big)^2,
    \quad m, M \in \real^{C},
\end{align}
where $b$ indexes the batch and $l$ indexes the spatial locations or tokens.
The two statistics are accumulated as a running average over the first 1000 training steps and then frozen.
Since $\phi$ is frozen, they converge quickly, and freezing them keeps the standardization fixed for the rest of training.
They are estimated on real features only and applied to both real and fake features.
This keeps the score models blind to the distribution difference between real and fake features, which is what the transport is meant to expose.

The standardization statistics of \cref{eq:score_matching_clean_feature} are then
\begin{align}
    \mu = m,
    \qquad
    s = \frac{\sigma_\mathrm{feat}}{\sigma_\mathrm{data}},
    \qquad
    \sigma_\mathrm{feat} = \sqrt{M - m^2} \in \real^{C},
\end{align}
so that $\bv = (\phi(\X) - \mu) / s$ has zero mean and standard deviation $\sigma_\mathrm{data}$ in every channel.
For ConvNeXt V2-N, we keep $s$ per-channel wise, because a few channels have far larger variance than the rest.
This is because the Global Response Normalization (GRN) layer leaves one channel dominating the others in magnitude.
Each channel's standard deviation is floored at $0.1$ times the median over channels, so a nearly constant channel is not amplified.
For DINO-S/8, the channels are naturally balanced, and we use a single scalar $s$ per layer, the root mean of the per-channel variances.

\subsection{Transport Direction}

The direction $\g$ in \cref{eq:score_matching_transport} is computed per sample.
Its denominator is the mean absolute denoising error of the real score model over the $LC$ entries of the feature map, floored at a small constant,
\begin{align}
    \label{eq:score_matching_transport_g}
    \g = \frac{
        \psi_\mathrm{fake}(\hat{\bv}_\sigma, \sigma) - \psi_\mathrm{real}(\hat{\bv}_\sigma, \sigma)
    }{n},
    \qquad
    n = \max\!\Big(
    \frac{1}{LC} \big\| \hat{\bv} - \psi_\mathrm{real}(\hat{\bv}_\sigma, \sigma) \big\|_1,\ 10^{-2}
    \Big).
\end{align}
The floor only guards against division by a vanishing error when a sample already lies on the real manifold at this $\sigma$; in practice it sits well below the typical value of $n$.
The noise level $\sigma$ is drawn from $q(\sigma)$ and clamped to $[0.002, 8.0]$.
Both score models are frozen here, so $\g$ carries no gradient.

\subsection{Transport Balance}

\cref{eq:score_matching_loss} is applied to every scored layer $j$ of $\phi$.
In practice we average the squared error over the $L_j C_j$ entries,
\begin{align}
    \cL_j = \frac{1}{2 L_j C_j} \| \hat{\bv}_j - \mathrm{sg}(\hat{\bv}_j - \g_j) \|^2_2,
    \qquad
    \frac{\partial \cL_j}{\partial \hat{\bv}_j} = \frac{1}{L_j C_j}\, \g_j ,
\end{align}
which differs from \cref{eq:score_matching_loss} by a constant $1/(L_j C_j)$ that the balance weights below absorb.
Since $\hat{\bv}_j$ is standardized with divisor $s_j = \sigma_{\mathrm{feat}}^j / \sigma_\mathrm{data}$ in \cref{eq:score_matching_clean_feature}, $\partial \hat{\bv}_j / \partial \phi_j = 1/s_j$, and the gradient reaching the generated image is
\begin{align}
    \frac{\partial \cL_j}{\partial \hat{\X}}
    = \frac{1}{L_j C_j}\, \frac{1}{s_j}\;
    \left( \frac{\partial \phi_j(\hat{\X})}{\partial \hat{\X}} \right)^{\!\top} \g_j .
\end{align}
Its magnitude therefore scales as
\begin{align}
    a_j \;\triangleq\; \frac{\| \g_j \|_2}{L_j C_j \; s_j},
    \qquad
    \big\| \partial \cL_j / \partial \hat{\X} \big\|_2 \le \kappa_j \, a_j,
    \quad
    \kappa_j = \big\| \partial \phi_j / \partial \hat{\X} \big\|_2 .
\end{align}
All three factors of $a_j$ depend on the layer properties: $L_j C_j$ is fixed, $s_j$ is known, and $\| \g_j \|_2$ is estimated on the fly.
For DINO-S/8, $s_j$ is the per-layer scalar of the feature standardization.
For ConvNeXt V2-N, whose $s_j$ is a per-channel vector, we collapse it to the scalar $\big( \tfrac{1}{C_j} \sum_c s_{j,c}^{-2} \big)^{-1/2}$.
This is the effective divisor: the gradient in channel $c$ is scaled by $1/s_{j,c}$, so its norm over channels is $\| \g_j \|_2$ times the root mean square of $1/s_{j,c}$, assuming $\g_j$ spreads evenly across channels.

Given a target magnitude $T$, we weight each layer by
\begin{align}
    w_j = \frac{T}{a_j},
    \qquad
    \cL_\mathrm{score} = \frac{1}{J} \sum_{j=1}^{J} w_j \, \cL_j ,
\end{align}
so that $w_j a_j = T$ for every layer, and the average over the $J$ scored layers keeps the loss weight independent of how many feature layers are used.
Because the regression target of \cref{eq:score_matching_transport_g} is detached in the proxy loss function, $\g_j$ does not depend on $w_j$, and scaling $\cL_j$ by $w_j$ scales its gradient by exactly $w_j$, so no custom backward is needed.

Since $\| \g_j \|_2$ drifts as the generator improves, we track it with a running mean over the first 1000 transport steps on Oxford Flowers and 10000 on ImageNet, and an EMA with decay $0.99$ afterwards.
The target $T$ is set to the mean of the $a_j$ during those 1000 steps and then frozen.
The balance thus equalizes the layers and, at the same time, holds the overall gradient magnitude at its calibrated level rather than letting it fade as the fake features approach the real ones.
$w_j$ is clipped to $[0.05, 20]$ as a guard against a collapsed layer taking unbounded weight.

\subsection{Training Dynamics of \texorpdfstring{$\cL_\mathrm{score}$}{L\_score}}
\label{appsec:score_dynamics}

\begin{figure}[!h]
    \centering
    \includegraphics[width=1.0\linewidth]{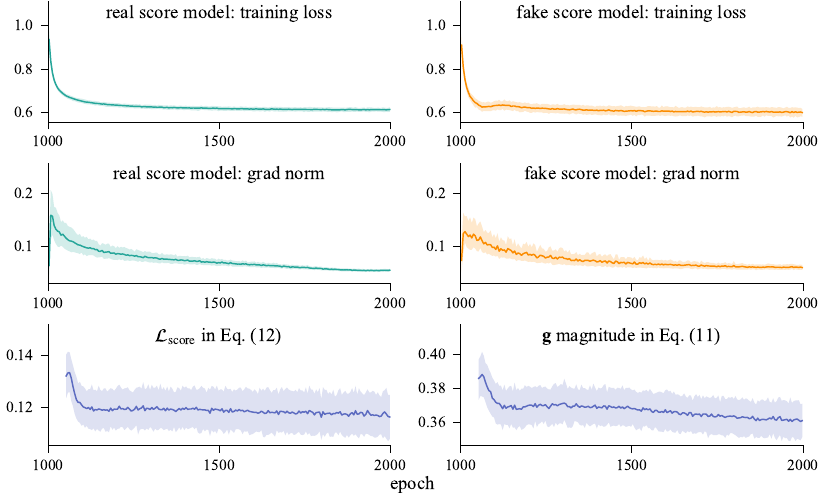}
    \caption{
        \textbf{Training dynamics of $\cL_\mathrm{score}$} for Sphere2-B on Oxford Flowers $256 \times 256$, from epoch 1000, where $\cL_\mathrm{score}$ is switched on, to epoch 2000.
        Top and middle rows: training loss and gradient norm of the two-layer real (left) and fake (right) score models, with a shared $y$ range per row.
        Bottom row: $\cL_\mathrm{score}$ in \cref{eq:score_matching_loss} and the magnitude of $\g$ in \cref{eq:score_matching_transport}.
        Each curve is the mean over 5 epochs, and the band spans the 10th to 90th percentiles.
    }
    \label{fig:score_curves}
\end{figure}

In \cref{fig:score_curves}, we show the training dynamics of $\cL_\mathrm{score}$ for Sphere2-B on Oxford Flowers $256 \times 256$, from epoch 1000, where $\cL_\mathrm{score}$ is switched on, to epoch 2000.
We found in the two-layer score models that QK-norm \citep{henry2020query} is essential for stable training.
Also, the hidden dimension of the score models should be at least the same as the feature dimension of the feature extractor $\phi$; otherwise, the score models are also unstable and diverge during the long training.

\section{Discussion and Limitations}

Beyond the conventional autoencoder objective $\cL_\mathrm{pix\text{-}recon}$ for training, Sphere Encoder 2 adds generation losses that are computed in frozen feature spaces, so decoder training keeps two extra networks in memory.
The first is DINOv3-small+ (29M), which serves as the frozen encoder $\cE$ and is shared as the feature extractor of $\cL_\mathrm{lat\text{-}con}$, adding one encoder forward on the generated image.
The second is ConvNeXt V2-N (14.98M), the feature extractor $\phi$ of $\cL_\mathrm{score}$; together with the real and fake score models of two transformer layers each, this loss module has 24-29M parameters in total.
$\cL_\mathrm{sem}$ needs no extra network, since it shares the LPIPS forward of $\cL_\mathrm{pix\text{-}recon}$.
This overhead applies only to training: sampling runs the encoder and decoder alone (\cref{alg:forward}).

\clearpage
\section{Implementation Details}
\label{appsec:impl}

\cref{tab:impl_config} lists the architecture, training, score matching loss, and sampling configurations of Sphere2-B and Sphere2-L used for the Oxford Flowers results in \cref{tab:fnl_flowers} and the ImageNet results in \cref{tab:fnl_imagenet}.
\cref{tab:fdr6_models} lists the six representation models used to compute FDr$^{6}$ at $256 \times 256$ and $512 \times 512$ on both Oxford Flowers and ImageNet.

\begin{table}[h]
    \centering
    \caption{
        \textbf{Configurations of experiments.}
        Values in a merged cell are shared by both models.
    }
    \label{tab:impl_config}
    \footnotesize
    \setlength{\tabcolsep}{5.0pt}
    \renewcommand{\arraystretch}{0.88}
    \newcolumntype{M}{>{\centering\arraybackslash}p{4.1cm}}
    \begin{tabular}{l M M}
        \toprule
                                                                 & Sphere2-B                                                                                           & Sphere2-L  \\
        \cmidrule(ll){2-3}
        \rowcolor{gray1}
        \multicolumn{3}{l}{\textit{architecture}}                                                                                                                                   \\
        encoder $\cE$                                            & \multicolumn{2}{c}{DINOv3-small+ (frozen, 29M)}                                                                  \\
        latent                                                   & \multicolumn{2}{c}{$N = 16 \times 16$, $D = 384$, $d = 98{,}304$}                                                \\
        decoder $\cD$                                            & ViT-B                                                                                               & ViT-L      \\
        depth                                                    & 12                                                                                                  & 24         \\
        hidden dim                                               & 768                                                                                                 & 1024       \\
        heads                                                    & 12                                                                                                  & 16         \\
        decoder attention qk norm                                & \multicolumn{2}{c}{True}                                                                                         \\
        decoder attention backend                                & \multicolumn{2}{c}{torch fp32 \texttt{SDPBackend.EFFICIENT\_ATTENTION}}                                          \\
        image size                                               & \multicolumn{2}{c}{256, 512}                                                                                     \\
        patch size of $\cD$                                      & \multicolumn{2}{c}{\texttt{image\_size} / 16}                                                                    \\
        conditioning                                             & \multicolumn{2}{c}{class $\y$ and angle $\alpha$ via adaLN-Zero}                                                 \\
        params (encoder + decoder)                               & 161M                                                                                                & 487M       \\
        GFLOPs (single forward)                                  & 22                                                                                                  & 115        \\
        \midrule
        \rowcolor{gray1}
        \multicolumn{3}{l}{\textit{training}}                                                                                                                                       \\
        precision                                                & \multicolumn{2}{c}{torch bf16 autocast}                                                                          \\
        epochs                                                   & \multicolumn{2}{c}{2000 on Oxford Flowers, 300 on ImageNet}                                                      \\
        warmup epochs                                            & \multicolumn{2}{c}{10}                                                                                           \\
        optimizer                                                & \multicolumn{2}{c}{AdamW}                                                                                        \\
        batch size                                               & \multicolumn{2}{c}{256}                                                                                          \\
        learning rate                                            & \multicolumn{2}{c}{$1 \times 10^{-4}$}                                                                           \\
        learning rate schedule                                   & \multicolumn{2}{c}{constant}                                                                                     \\
        weight decay                                             & \multicolumn{2}{c}{0}                                                                                            \\
        ema decay                                                & \multicolumn{2}{c}{0.99985}                                                                                      \\
        angle sampler                                            & \multicolumn{2}{c}{$\alpha \sim \mathcal{U}[0^\circ, 90^\circ]$}                                                 \\
        cutoff angle $\alpha_\mathrm{cutoff}$, bandwidth $\beta$ & \multicolumn{2}{c}{$70^\circ$, $5^\circ$}                                                                        \\
        data augmentation                                        & \multicolumn{2}{c}{center crop, horizontal flip $p = 0.5$}                                                       \\
        feature extractors for $\cL_\mathrm{FD\text{-}lite}$     & \multicolumn{2}{c}{Inception, DINOv3-small+, ConvNeXt V2-Nano}                                                   \\
        class dropout probability (for CFG)                      & \multicolumn{2}{c}{0.1}                                                                                          \\
        attention residual dropout probability                   & \multicolumn{2}{c}{0.1}                                                                                          \\
        \midrule
        \rowcolor{gray1}
        \multicolumn{3}{l}{\textit{score matching loss}}                                                                                                                            \\
        training start epoch                                     & \multicolumn{2}{c}{1000 on Oxford Flowers, 100 on ImageNet}                                                      \\
        transport start epoch                                    & \multicolumn{2}{c}{1150 on Oxford Flowers, 115 on ImageNet}                                                      \\
        score feature space $\phi$                               & \multicolumn{2}{c}{ConvNeXt V2-N \texttt{ft-in1k} (frozen, 16M), two middle stages}                              \\
        score models $\psi_\mathrm{real}, \psi_\mathrm{fake}$    & \multicolumn{2}{c}{2-layer transformers, EDM preconditioning}                                                    \\
        score model conditioning                                 & \multicolumn{2}{c}{class $\y$ via adaLN-Zero}                                                                    \\
        score model hidden dim                                   & \multicolumn{2}{c}{384}                                                                                          \\
        score model attention qk norm                            & \multicolumn{2}{c}{True}                                                                                         \\
        score model attention backend                            & \multicolumn{2}{c}{torch bf16 \texttt{SDPBackend.EFFICIENT\_ATTENTION}}                                          \\
        score model learning rate                                & \multicolumn{2}{c}{$5 \times 10^{-4}$, same for $\psi_\mathrm{real}$ and $\psi_\mathrm{fake}$}                   \\
        score model learning rate schedule                       & \multicolumn{2}{c}{cosine decay to $5 \times 10^{-5}$}                                                           \\
        score $\sigma$ for training $p(\sigma)$                  & \multicolumn{2}{c}{log-normal $(p_\mathrm{mean}, p_\mathrm{std}) = (-1.2, 1.2)$, no clip}                        \\
        score $\sigma$ for transport $q(\sigma)$                 & \multicolumn{2}{c}{log-normal $(q_\mathrm{mean}, q_\mathrm{std}) = (-1.5, 1.0)$, clip $[0.002, 8]$}              \\
        transport balance                                        & \multicolumn{2}{c}{EMA $0.99$, $w_j \in [0.05, 20]$}                                                             \\
        differentiable augmentation                              & \multicolumn{2}{c}{horizontal flip $p = 0.5$, translate $p = 1.0$ with pixel ratio $1/64$}                       \\
        params (all score models)                                & \multicolumn{2}{c}{24$\sim$28M}                                                                                  \\
        \midrule
        \rowcolor{gray1}
        \multicolumn{3}{l}{\textit{loss weights}}                                                                                                                                   \\
        $\cL_\mathrm{pix\text{-}recon}$                          & \multicolumn{2}{c}{1.0}                                                                                          \\
        $\cL_\mathrm{sem}$                                       & \multicolumn{2}{c}{1.0}                                                                                          \\
        $\cL_\mathrm{score}$                                     & \multicolumn{2}{c}{1.0}                                                                                          \\
        $\cL_\mathrm{lat\text{-}con}$                            & \multicolumn{2}{c}{0.1}                                                                                          \\
        $\cL_\mathrm{FD\text{-}lite}$                            & \multicolumn{2}{c}{0.1 for Sphere2$^\dagger$}                                                                    \\
        \midrule
        \rowcolor{gray1}
        \multicolumn{3}{l}{\textit{sampling}}                                                                                                                                       \\
        sampling schedule                                        & \multicolumn{2}{c}{loop with constant $\alpha$}                                                                  \\
        sampling angle $\alpha$                                  & $85.5^\circ$                                                                                        & $84^\circ$ \\
        guidance angle $\gamma$                                  & $15^\circ$                                                                                          & $5^\circ$  \\
        \bottomrule
    \end{tabular}
\end{table}

\clearpage
\begin{table}[h]
    \centering
    \caption{
        \textbf{Representation models of FDr$^{6}$} at $256 \times 256$ and $512 \times 512$.
        The $256 \times 256$ set follows the setup of FDr$^{6}$ \citep{yang2026representation}, which is originally for ImageNet $256 \times 256$.
        At $512 \times 512$, each family uses the highest-resolution checkpoint.
        Inception-v3 and MAE have no such checkpoint and keep the ones used at $256 \times 256$.
    }
    \label{tab:fdr6_models}
    \footnotesize
    \setlength{\tabcolsep}{4pt}
    \resizebox{\linewidth}{!}{%
        \begin{tabular}{l l c c c c c}
            \toprule
            Model                                      & timm identifier                                       & Arch. & Dim  & Input & Objective       & Pooling         \\
            \midrule
            \rowcolor{gray1}
            \multicolumn{7}{l}{\textit{Image Size $256 \times 256$}}                                                                                                      \\
            Inception-v3 \citep{szegedy2016rethinking} & \texttt{inception\_v3} (torch-fidelity)               & CNN   & 2048 & 299   & supervised      & global avg pool \\
            ConvNeXt-v2 \citep{woo2023convnext}        & \texttt{convnextv2\_base.fcmae\_ft\_in22k\_in1k}      & CNN   & 1024 & 224   & self-supervised & global avg pool \\
            MAE \citep{he2022masked}                   & \texttt{vit\_large\_patch16\_224.mae}                 & ViT   & 1024 & 224   & reconstructive  & CLS token       \\
            DINOv2 \citep{oquab2023dinov2}             & \texttt{vit\_large\_patch14\_dinov2.lvd142m}          & ViT   & 1024 & 256   & contrastive     & CLS token       \\
            SigLIP2 \citep{tschannen2025siglip}        & \texttt{vit\_so400m\_patch16\_siglip\_256.v2\_webli}  & ViT   & 1152 & 224   & vision-language & CLS token       \\
            CLIP \citep{radford2021learning}           & \texttt{vit\_large\_patch14\_clip\_224.openai}        & ViT   & 1024 & 256   & vision-language & CLS token       \\
            \midrule
            \rowcolor{purple0}
            \multicolumn{7}{l}{\textit{Image Size $512 \times 512$}}                                                                                                      \\
            Inception-v3 \citep{szegedy2016rethinking} & \texttt{inception\_v3} (torch-fidelity)               & CNN   & 2048 & 299   & supervised      & global avg pool \\
            ConvNeXt-v2 \citep{woo2023convnext}        & \texttt{convnextv2\_base.fcmae\_ft\_in22k\_in1k\_384} & CNN   & 1024 & 384   & self-supervised & global avg pool \\
            MAE \citep{he2022masked}                   & \texttt{vit\_large\_patch16\_224.mae}                 & ViT   & 1024 & 224   & reconstructive  & CLS token       \\
            DINOv2 \citep{oquab2023dinov2}             & \texttt{vit\_large\_patch14\_dinov2.lvd142m}          & ViT   & 1024 & 518   & contrastive     & CLS token       \\
            SigLIP2 \citep{tschannen2025siglip}        & \texttt{vit\_so400m\_patch16\_siglip\_512.v2\_webli}  & ViT   & 1152 & 512   & vision-language & CLS token       \\
            CLIP \citep{radford2021learning}           & \texttt{vit\_large\_patch14\_clip\_336.openai}        & ViT   & 1024 & 336   & vision-language & CLS token       \\
            \bottomrule
        \end{tabular}}
\end{table}

\bibliography{iclr2027}
\bibliographystyle{iclr2027}

\end{document}